\documentclass[final,3p,times,twocolumn,nopreprintline]{elsarticle} 

\usepackage{amssymb} 
\usepackage[normalem]{ulem} 
\useunder{\uline} {\ul} {} 
\usepackage{amsmath,amsfonts}
\usepackage{booktabs} 
\usepackage[table,xcdraw,dvipsnames]{xcolor} 
\usepackage{multirow} 
\usepackage{subfigure} 
\usepackage{url} 
\usepackage{array} 
\usepackage{caption} 
\usepackage{fancyhdr}
\newcommand{\red}[1]{\textcolor{Red}{#1}}
\newcommand{\green}[1]{\textcolor{ForestGreen}{#1}}
\usepackage{enumitem}
\usepackage{bbm}
\usepackage{tabularx}

\usepackage{algorithm}
\usepackage{algpseudocode}

\newcolumntype{L} [1]{>{\raggedright\let\newline\\\arraybackslash\hspace{0pt} } m{#1} } 
\newcolumntype{C} [1]{>{\centering\let\newline\\\arraybackslash\hspace{0pt} } m{#1} } 
\newcolumntype{R} [1]{>{\raggedleft\let\newline\\\arraybackslash\hspace{0pt} } m{#1} }

\definecolor{softsky}{RGB}{122, 150, 153}
\definecolor{softgray}{gray}{0.2}

\usepackage[colorlinks=true, linkcolor=blue, urlcolor=black, citecolor=blue, filecolor=blue]{hyperref}
\hypersetup{pdfauthor=author}

\journal{} 

\begin{document} 

\begin{frontmatter} 

\title{Leveraging Out-of-Distribution Unlabeled Images:\\ Semi-Supervised Semantic Segmentation with an Open-Vocabulary Model} 

\author{Wooseok Shin} 
\ead{wsshin95@korea.ac.kr}
\author{Jisu Kang} 
\ead{ji_soo_o@korea.ac.kr}
\author{Hyeonki Jeong} 
\ead{gusrl1210@korea.ac.kr}
\author{Jin Sob Kim}
\ead{jinsob@korea.ac.kr}
\author{Sung Won Han\corref{cor1}} 
\ead{swhan@korea.ac.kr}
\cortext[cor1]{Corresponding author} 
\address{Department of Industrial and Management Engineering, Korea University, Seoul, Republic of Korea} 

\begin{abstract} 
In semi-supervised semantic segmentation, existing studies have shown promising results in academic settings with controlled splits of benchmark datasets. However, the potential benefits of leveraging significantly larger sets of unlabeled images remain unexplored.
In real-world scenarios, abundant unlabeled images are often available from online sources (web-scraped images) or large-scale datasets.
However, these images may have different distributions from those of the target dataset, a situation known as out-of-distribution (OOD). Using these images as unlabeled data in semi-supervised learning can lead to inaccurate pseudo-labels, potentially misguiding network training.
In this paper, we propose a new semi-supervised semantic segmentation framework with an open-vocabulary segmentation model (SemiOVS) to effectively utilize unlabeled OOD images.
Extensive experiments on Pascal VOC and Context datasets demonstrate two key findings: (1) using additional unlabeled images improves the performance of semi-supervised learners in scenarios with few labels, and (2) using the open-vocabulary segmentation (OVS) model to pseudo-label OOD images leads to substantial performance gains.
In particular, SemiOVS outperforms existing PrevMatch and SemiVL methods by +3.5 and +3.0 mIoU, respectively, on Pascal VOC with a 92-label setting, achieving state-of-the-art performance. These findings demonstrate that our approach effectively utilizes abundant unlabeled OOD images for semantic segmentation tasks. We hope this work can inspire future research and real-world applications. The code is available at \href{https://github.com/wooseok-shin/SemiOVS}{https://github.com/wooseok-shin/SemiOVS}
\end{abstract}


\end{frontmatter} 

\section{Introduction}
Semantic segmentation plays a crucial role in numerous computer vision applications, including autonomous driving \cite{wang2020deep}, image editing \cite{ling2021editgan}, medical-image analysis \cite{shin2022comma}, robotics \cite{milioto2019bonnet}, and surveillance systems \cite{gruosso2021human}, where the objective is to assign a semantic class to each pixel within an image. Although approaches based on supervised learning have recently achieved significant success in this area, obtaining accurate pixel-level ground truths for supervised learning is time consuming and costly. This poses a challenge to the applicability of supervised-learning methods in various fields.
To mitigate this limitation, many researchers have focused on developing semi-supervised semantic segmentation methods, which leverage a small number of labeled images along with a large number of unlabeled images.

\begin{figure*}[ht!]
\centering
\includegraphics[width=1.0\linewidth]{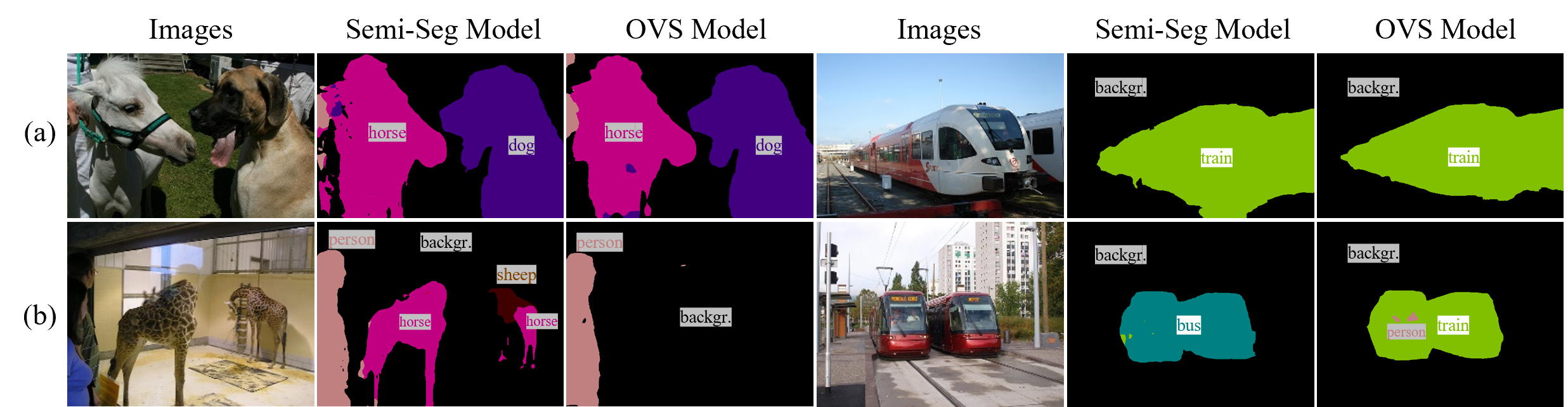}
\caption{Visualization of semantic segmentation results: (a) in-distribution images and pseudo-labels on Pascal VOC, (b) OOD images and pseudo-labels on COCO. The mean intersection-over-union (mIoU) scores of the Semi-Seg and open-vocabulary segmentation (OVS) models on the Pascal VOC validation set are 79.0 and 78.9, respectively. The Semi-Seg Model shows accurate predictions for in-distribution samples but struggles with OOD images. For example, it misclassifies a giraffe as a horse or sheep instead of a background, and a light rail (train) as a bus. In contrast, the OVS model exhibits robust performance on both in-distribution and OOD samples, providing reliable pseudo-labels for images from both distributions.}
\label{fig:intro}
\end{figure*}

Existing semi-supervised semantic segmentation studies \cite{sohn2020fixmatch,yang2023revisiting,shin2024revisiting,hoyer2023semivl,chen2021semi,li2023diverse,wang2022semi} have shown promising results in academic settings where benchmark datasets are split into various setups based on different proportions or numbers of labeled and unlabeled images.
However, there has been little exploration of the potential benefits of leveraging more unlabeled images. In real-world scenarios, it would be more practical and beneficial to leverage substantially larger sets of unlabeled images, potentially improving model performance beyond what can be achieved in controlled academic settings.
Therefore, in this study, we hypothesize that using a significantly larger set of unlabeled images will improve the performance of a semi-supervised learner. By testing this hypothesis, we aim to demonstrate a practical and effective solution for real-world applications, where unlabeled data is often abundant but labeled data is scarce.

Regarding the sources of additional unlabeled images, two primary approaches can be considered.
The first approach involves collecting in-distribution unlabeled images, that is, images whose distribution closely matches that of the target dataset. For example, if the target dataset is Pascal VOC \cite{everingham2010pascal}, additional images from a similar distribution would need to be gathered. However, this approach can be time consuming and costly as it requires the acquisition of data specifically tailored to the distribution of interest.

Alternatively, unlabeled images can be obtained from readily available sources such as websites (web-scraped images) or large-scale datasets~\footnote{Note that while large-scale datasets may include annotations, only the images were used in this study to investigate the impact of additional unlabeled data on performance improvement.} (e.g., COCO \cite{lin2014microsoft}). These datasets often share certain domain characteristics with the target dataset, such as common object classes. For example, the 20 object classes Pascal VOC are frequently present in COCO or web-scraped images as well. This overlap allows for easier and more cost-effective large-scale image collection.
However, it is important to note that the unlabeled images acquired from these sources may have different distributions from that of the target dataset, a situation known as out-of-distribution (OOD). This discrepancy can manifest in various ways, such as variability in image quality and resolution, object poses and sizes, lighting conditions, and background complexity. These differences can result in inaccurate pseudo-labels when OOD images are used as unlabeled data in a semi-supervised learning process.

Fig.~\ref{fig:intro} illustrates these challenges; specifically, how a semantic segmentation model performs well on in-distribution samples but struggles with unseen objects in OOD scenarios.  For example, the standard segmentation model mistakenly classifies a giraffe as a horse or sheep owing to their similar shapes, rather than identifying it as part of the background. Similarly, while the model accurately segments typical trains, it fails to classify newer types of trains, such as light rail, misclassifying them as buses. These cases show that naively leveraging OOD images can mislead network training.

In this paper, we propose a new semi-supervised segmentation framework with an open-vocabulary segmentation model (SemiOVS) to effectively utilize unlabeled OOD images. The OVS model \cite{zhou2022extract,ghiasi2022scaling,xie2024sed}, pretrained on extensive image-text datasets \cite{radford2021learning}, can segment objects beyond the fixed categories of the labeled dataset using text descriptions and demonstrates strong generalizability. Therefore, to mitigate the aforementioned problem involving distribution shifts, we integrate an OVS model into the existing semi-supervised learning process. 
In particular, the OVS model generates pseudo-labels for OOD images. Then, the standard segmentation model uses these pseudo-labels to learn OOD objects. This strategy provides the standard segmentation model with reliable guidance for OOD images, expanding its understanding to objects and scenes beyond the in-distribution data.

\begin{figure}[t!]
\centering
\includegraphics[width=1.0\linewidth]{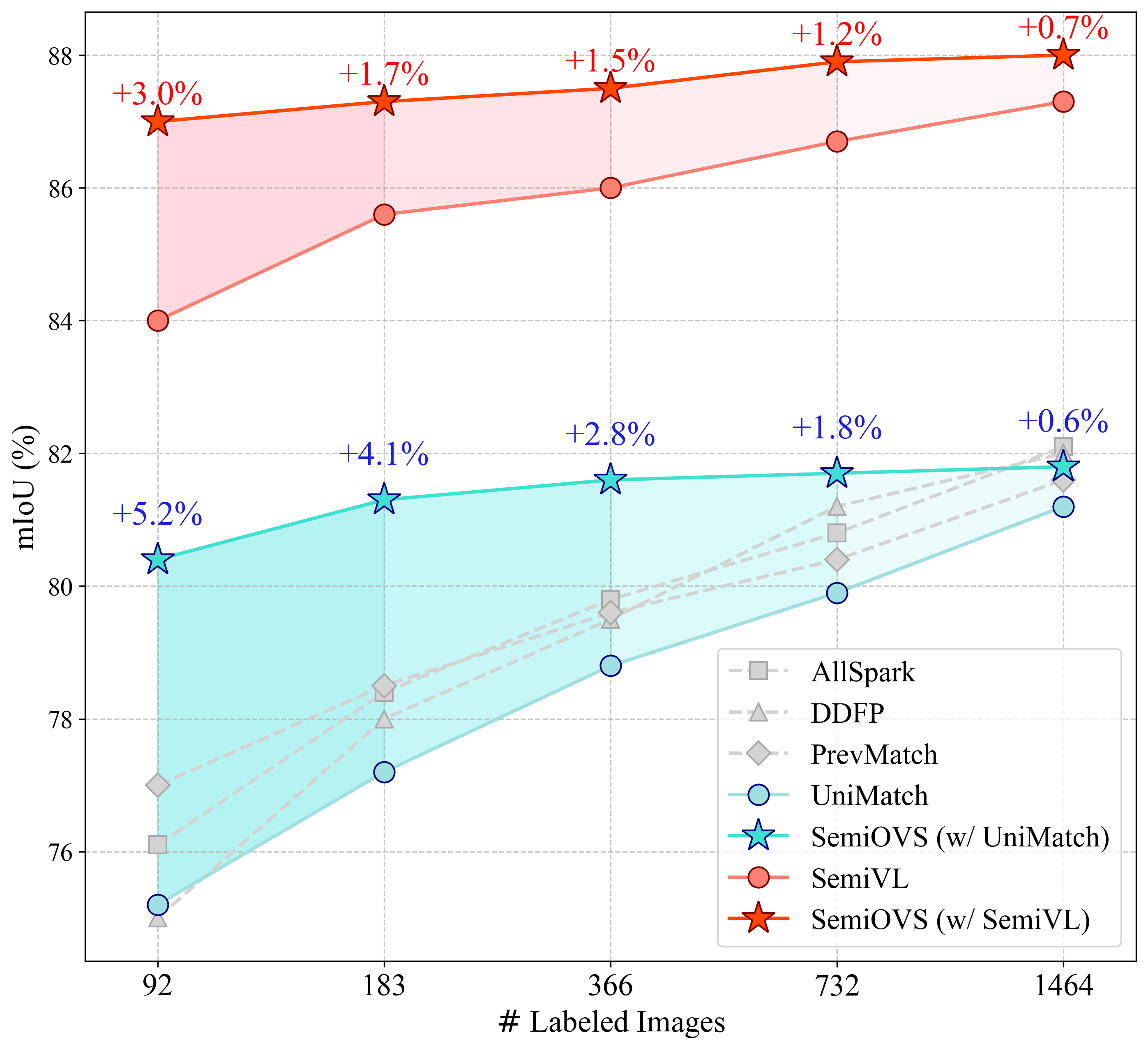}
\caption{Comparison with state-of-the-art methods on the Pascal VOC \textit{Original} protocol.}
\label{fig:comparison}
\end{figure}

Extensive experiments on the Pascal VOC and Pascal Context datasets reveal that (1) leveraging additional unlabeled images from the COCO dataset or online sources significantly improves the performance of the semi-supervised learner, and (2) using the OVS model to pseudo-label OOD images substantially improves performance.
Moreover, the proposed SemiOVS achieves state-of-the-art performance across various evaluation protocols, as shown in Fig.~\ref{fig:comparison}, without incurring additional computational costs during inference.
Under ResNet-50 and DeepLabV3+ network settings, SemiOVS outperforms the existing state-of-the-art method, PrevMatch, by +4.8 mIoU on the Pascal VOC dataset, with a 92-label setting.
Finally, SemiOVS is compatible with existing semi-supervised segmentation methods, enabling easy integration into various frameworks.

Our contributions are as follows:
\begin{itemize}
    \setlength{\itemindent}{-4pt}
    \item We investigate the efficacy of using additional unlabeled images in semi-supervised semantic segmentation, offering a more practical solution for real-world applications where large amounts of unlabeled data are available.
    \item To mitigate the problem of unlabeled OOD image distribution discrepancy in pseudo-labeling, we propose SemiOVS, a novel framework that integrates an OVS model, augmented by two key components: (1) a text prompt set including potential OOD classes and (2) a refinement process aligning initial pseudo-labels to the target label space.
    \item The proposed SemiOVS outperforms existing state-of-the-art methods by a large margin across different evaluation protocols without incurring additional computational costs during inference.
\end{itemize}

\section{Related Work}
\subsection{Semi-supervised Learning}
The primary objective of semi-supervised learning, including semi-supervised segmentation, is to extract meaningful information from unlabeled samples, thereby enhancing model generalizability.
To this end, consistency regularization and self-training have been widely used as dominant paradigms.
Consistency regularization \cite{sohn2020fixmatch,laine2016temporal,tarvainen2017mean} encourages a model to maintain consistent predictions for different perturbations of the same input, thus improving its robustness to variations.
Self-training \cite{lee2013pseudo,arazo2020pseudo,yang2022st++}, also known as pseudo-labeling, operates by assigning pseudo-labels based on the model's own predictions for unlabeled samples. These pseudo-labels are then used to train the model alongside labeled data, progressively refining its understanding of the underlying data distribution.

Recent studies have combined these two paradigms to design more robust and effective frameworks.
Representatively, FixMatch introduces a weak-to-strong consistency regularization method. In particular, FixMatch \cite{sohn2020fixmatch} applies two levels of perturbations to unlabeled images: weak and strong. The model then generates pseudo-labels from the prediction for the weak perturbation and uses them to supervise the prediction for the strong perturbation.
The core concept behind this approach is that pseudo-labels derived from weak perturbations are generally more reliable, while strong perturbations help mitigate confirmation bias and broaden the understanding of the model.
Based on the success of FixMatch, recent semi-supervised semantic segmentation studies have adopted the weak-to-strong consistency paradigm as a fundamental component.

\subsection{Semi-supervised Semantic Segmentation}
Since semantic segmentation involves various contexts with multiple classes within an image, it requires advanced approaches for accurate performance. Therefore, recent studies have designed sophisticated frameworks for semi-supervised segmentation, building on the advances made in semi-supervised learning.
One popular approach is the teacher-student framework \cite{tarvainen2017mean}, where a teacher model is derived as an exponential moving average (EMA) of the student model. Based on this approach, existing studies have developed diverse strategies including CutMix-based augmentation \cite{french2019semi, hu2021semi}, contrastive learning \cite{lai2021semi,wang2022semi,wang2023hunting}, prototype learning \cite{xu2022semi}, symbolic reasoning \cite{liang2023logic}, and multiple EMA teachers \cite{liu2022perturbed, na2023switching}.
Another popular strategy is co-training \cite{qiao2018deep} where two differently initialized networks are trained to mutually supervise each other, promoting diversity in learning.
Studies based on this approach have proposed various strategies such as cross-pseudo supervision \cite{chen2021semi}, shared encoder with multiple final heads \cite{fan2022ucc}, enhancing co-training diversity \cite{li2023diverse}, and conflict-based pseudo-labeling \cite{wang2023conflict}.

Instead of using multiple networks, which complicates training, UniMatch \cite{yang2023revisiting} expands the concept of FixMatch \cite{sohn2020fixmatch} via feature-level and dual-branch image-level perturbations.
Subsequently, CorrMatch \cite{sun2024corrmatch} introduces a label propagation strategy with correlation matching to find more reliable regions, and PrevMatch \cite{shin2024revisiting} uses temporal knowledge obtained during training to mitigate the confirmation bias problem. Moreover, several studies \cite{hoyer2023semivl,huang2023semicvt,wang2024allspark} have introduced transformer-based structures in semi-supervised semantic segmentation. 
Among them, Hoyer et al. proposed SemiVL \cite{hoyer2023semivl}, which is the first method to leverage vision-language models in semi-supervised semantic segmentation.
Unlike the aforementioned studies, we explore the impact of leveraging additional unlabeled images, an aspect not addressed in previous research.

\subsection{Open-vocabulary Semantic Segmentation}
Open-vocabulary semantic segmentation involves assigning a semantic class label to each pixel in an image using arbitrary text queries, without being limited to a predefined set of classes. To this end, open-vocabulary semantic segmentation requires precise alignment between visual and textual features in a joint embedding space.
Recently, vision-language models have been developed to align and integrate visual and linguistic representations, enabling a more seamless connection between these two domains. One prominent method is CLIP \cite{radford2021learning}, which leverages a massive dataset of 400 million image-text pairs collected from web sources. It uses contrastive learning to align images and texts in a shared embedding space.
This enables open-vocabulary or zero-shot recognition, which classifies various visual concepts using natural language descriptions, even for categories not seen during training.

Researchers have extended the application of CLIP to semantic segmentation, investigating how the learned representations of CLIP can be leveraged for pixel-level classification tasks.
Several studies \cite{zhou2022extract,cha2023learning,shin2022reco} have explored zero-shot approaches, which focus on learning segmentation solely from image-caption pairs, without requiring pixel-level annotations.
In particular, MaskCLIP \cite{zhou2022extract} obtains pixel-level feature maps by slightly modifying the CLIP architecture, removing the self-attention pooling layer. It then produces the final segmentation mask by combining these fine-grained visual features with text embeddings.
Although these methods perform reasonably, the predicted masks tend to be noisy owing to the lack of dense-level supervision.
To address this issue, subsequent studies \cite{ghiasi2022scaling,xu2022simple,liang2023open,xu2023side,yu2024convolutions,cho2024cat,xie2024sed} incorporated additional dense-level annotations to train segmentation networks or fine-tuning the CLIP encoder, where CLIP is used to segment unseen classes not included in the dense-level annotations.

In this study, we integrate the OVS method into the existing semi-supervised segmentation framework to mitigate the distribution discrepancy that arises when using OOD images as unlabeled data.
This offers two main advantages: (1) it enables the effective use of OOD images as unlabeled data, reducing the need for strictly in-distribution data collection and thereby simplifying the data acquisition process, and (2) it improves model generalizability by exposing the model to a wider range of visual cases.
Consequently, we provide a practical solution to yield robust segmentation models under scenarios where labeled data is scarce.

\begin{figure*}[t]
\centering
\includegraphics[width=0.96\linewidth]{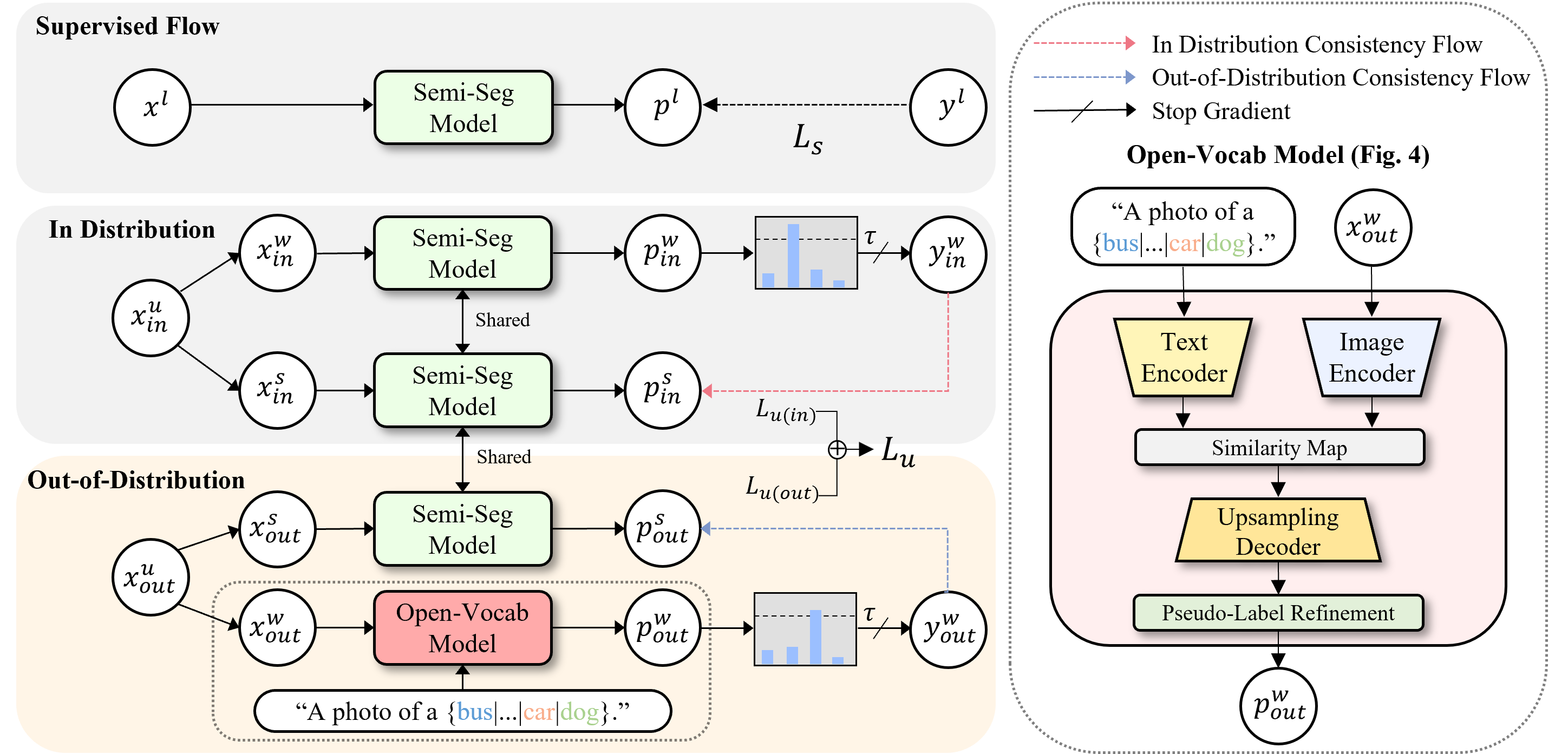}
\caption{The overall framework of SemiOVS, which consists of a standard semi-supervised segmentation model and an OVS model. In-distribution unlabeled images are pseudo-labeled using the standard segmentation model, while unlabeled OOD images are handled via the OVS model.}
\label{fig:framework}
\end{figure*}

\section{Methodology}

\subsection{Preliminaries and Overall Workflow}
\label{subsec3:preliminary}
The goal of semi-supervised semantic segmentation is to maximize the utility of unlabeled images $D_u = \{x^{u}_{i}\}$, given that only a small number of labeled images $D_l = \{(x^{l}_{i}, y^{l}_{i})\}$ are available.
In semi-supervised learning, the objective function typically comprises two components: a supervised loss term $L_s$ and an unsupervised loss term $L_u$, as shown below:
\begin{equation}
\label{eq1}
L = L_s + L_u.
\end{equation}

\begin{enumerate}[label=\arabic*)]
    \item \textbf{Supervised Flow.}
    In the supervised flow, a segmentation model $f$ takes a labeled image $x^{l}$ as input and outputs the prediction $p^{l}$. The supervised loss is then computed using pixel-level cross-entropy $H$ between the model prediction and the ground-truth annotation. This can be mathematically expressed as follows.
    \begin{equation}
      L_s = \sum H(y^l, p^l).
      \label{eq2}
    \end{equation}

    \item \textbf{Unsupervised -- In-Distribution Flow.}
    Most semi-supervised segmentation approaches have adopted the weak-strong consistency regularization paradigm to generate supervision for unlabeled images, a technique that gained prominence through FixMatch \cite{sohn2020fixmatch}. Specifically, an unlabeled image $x^u_{in}$ is subjected to two types of perturbations: weak and strong perturbations, resulting in $x^w_{in}$ and $x^s_{in}$ respectively. These perturbed images are then fed into the model $f$, which produces the predicted class distributions $p^w_{in}$ and $p^s_{in}$.
    Subsequently, the strongly perturbed prediction $p^s_{in}$ is supervised using the pseudo-label derived from the weakly augmented prediction $p^w_{in}$, obtained as $y^w_{in} = \text{argmax}(p^w_{in})$, as shown in the in-distribution flow in Fig.~\ref{fig:framework} (red arrow). This consistency regularization term can be formulated as follows.
    \begin{equation}
    \label{eq3}
    L_{u(in)} = \sum\mathbbm{1}(max(p^w_{in}) \geq \tau) \; H(y^w_{in}, p^s_{in}).
    \end{equation}
    Here, $\mathbbm{1}(\cdot)$ represents an indicator function, and $\tau$ denotes the confidence threshold used to filter out noise in the pseudo-label.

    \item \textbf{Unsupervised -- Out-of-Distribution Flow.}
    First, we assume that there are unlabeled images $x^u_{out}$ scraped from web sources or obtained from public datasets such as COCO. Two perturbed OOD images, $x^w_{out}$ and $x^s_{out}$, are then generated by applying weak and strong perturbations, respectively, to the unlabeled OOD image $x^u_{out}$.
    The OOD flow follows a similar structure to the in-distribution flow, but with key differences to address the challenges caused by the distribution shift. Specifically, the strongly augmented image $x^s_{out}$ is fed into the segmentation model $f$ to obtain $p^s_{out}$, whereas the weakly augmented image $x^w_{out}$ is processed by a pretrained OVS model $g$, yielding the prediction $p^w_{out}$ and the corresponding pseudo-label $y^w_{out}$, where $y^w_{out}$ is obtained as $y^w_{out} = \text{argmax}(p^w_{out})$ Then, $p^s_{out}$ is supervised using $y^w_{out}$, as depicted in the OOD flow in Fig.~\ref{fig:framework} (blue arrow). The term for OOD can be formulated as follows.    
    \begin{equation}
    \label{eq4}
    L_{u(out)} = \sum\mathbbm{1}(max(p^w_{out}) \geq \tau) \; H(y^w_{out}, p^s_{out}).
    \end{equation}

\end{enumerate}
Finally, the total unsupervised loss is computed as follows:
\begin{equation}
    L_u = L_{u(in)} + \lambda \cdot L_{u(out),}
  \label{eq5}
\end{equation}
where $\lambda$ indicates the weight of the OOD flow.

\begin{figure*}[t]
\centering
\includegraphics[width=0.96\linewidth]{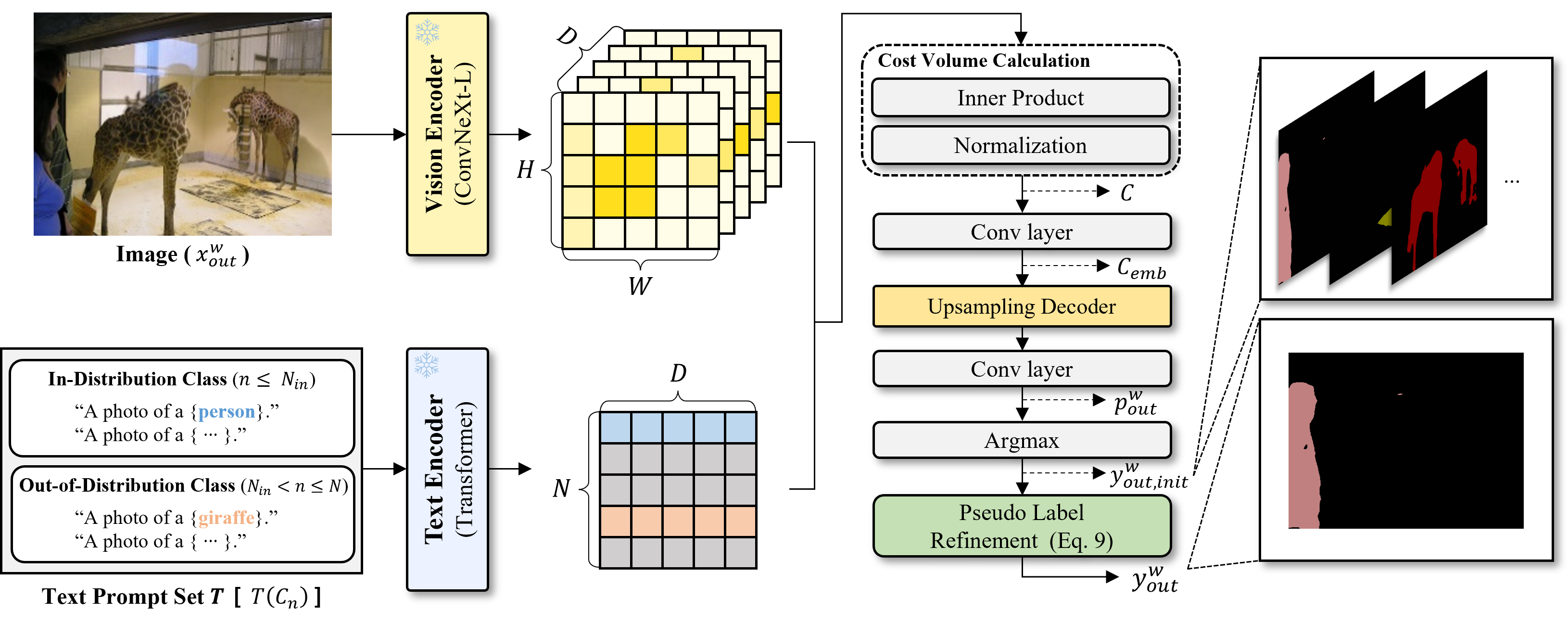}
\caption{Detailed illustration of the OVS process within the OOD flow. An unlabeled OOD image and a set of text prompts are processed through a pretrained vision-language model to generate image and text embeddings. A similarity map is then computed between the two embeddings, and decoding is applied to produce initial pseudo labels. These are further refined to obtain the final pseudo labels used for training in the target task.}
\label{fig:ovs_detail}
\end{figure*}

\subsection{Semi-supervised Semantic Segmentation with Open-vocabulary Model}
\label{subsec:3.2}
To mitigate the generation of incorrect pseudo-labels due to distribution shift for OOD images, we propose SemiOVS, which consists of a semi-supervised segmentation model ($f$) and an OVS model ($g$).
As depicted in Fig.~\ref{fig:framework}, in the in-distribution flow, both perturbed images ($x^w_{in}$ and $x^s_{in}$) are passed through the same segmentation model $f$. 
In contrast, in the OOD flow, $x^s_{out}$ is fed into the segmentation model $f$, while $x^w_{out}$ and the text prompt set $T$=\{$T(C_n)\}$ for $n$=$1,...,N$ are processed by the OVS model $g$, yielding $p^w_{out}$. Here, $T(C_n)$ is the text description for the $n$-th class $C_n$ using a prompt template, such as ``a photo of a $\{C_n\}$''. $N$ represents the total number of classes included in the text prompt set. This can be formulated as follows.
\begin{equation}
\label{eq6}
p^w_{out} = g(x^w_{out}, T), \qquad p^s_{out} = f(x^s_{out}).
\end{equation}
We now provide a detailed explanation of the OVS process.
As shown in Fig.~\ref{fig:ovs_detail}, the OVS model ($g$) consists of an image encoder and text encoder, defined as $g_v$ and $g_t$, respectively.
For the image encoding branch, the image encoder $g_v$ receives a weakly perturbed OOD image $x^w_{out}$ and extracts image embeddings $E_V$.
For the text encoding branch, given a set of class names $\{C_1, C_2, ..., C_N\}$, the text prompt set $T$=$\{T(C_1), T(C_2), ..., T(C_N)\}$ are generated for $N$ classes.
Subsequently, the text encoder processes $T$ and outputs text embeddings $E_T$. This can be formulated as:
\begin{equation}
\label{eq7}
E_V = g_v(x^w_{out}) \in \mathbb{R}^{H \times W \times D}, \; E_T = g_t(T) \in \mathbb{R}^{N \times D},
\end{equation}
where $D$ denotes the feature dimension.
Subsequently, a pixel-level cost volume $C \in \mathbb{R}^{H \times W \times N \times 1}$ (i.e., similarity matrix) is obtained by calculating the cosine similarity between image embeddings $E_V$ and text embeddings $E_T$ as follows:
\begin{equation}
\label{eq8}
C(i, j, n) = \frac{E_V(i,j) \cdot E_T(n)}{\lVert E_V(i,j) \rVert \; \lVert E_T(n) \rVert},
\end{equation}
where $i$,$j$ denotes the 2D spatial locations of the image embedding, and $n$ represents the class index for the text embedding. Then, the cost volume embedding $C_{emb} \in \mathbb{R}^{H \times W \times N \times D}$, which is used as the input features of the decoder $g_d$, is obtained by applying a convolutional layer to the initial cost volume $C$.
To generate high-resolution segmentation maps, it is necessary to enhance the low-resolution and noisy cost volume embedding $C_{emb}$. Therefore, similar to feature pyramid networks (FPN \cite{lin2017feature}), a decoder architecture is adopted that progressively upsamples the embedding while incorporating intermediate-level image features from the encoder.
Formally, $C_{emb}$ is processed through the decoder $g_d$ and final convolution layers to produce $p^w_{out} \in \mathbb{R}^{H \times W \times N}$. Subsequently, an initial pseudo-label $y^w_{out, init} \in \mathbb{R}^{H \times W \times 1}$ is obtained by applying $argmax$ operation to $p^w_{out}$.

The initial pseudo-labels can serve as supervisory signals in training the semi-supervised segmentation model. However, we introduce two key enhancements to the standard OVS process to effectively adapt them to the semi-supervised setting.
First, we construct an extended text prompt set that includes the predefined in-distribution classes (e.g., Pascal VOC 20 classes) and additional categories likely to appear in the unlabeled OOD images. This design allows the OVS model to produce more accurate pseudo-labels for diverse image sources (refer to Section \ref{subsec:prompt_set} for related analysis).  
We define the prompt set as $\{T(C_1),...,T(C_{N_{in}}), ..., T(C_N)\}$, where $N_{in} < N$. Consequently, each pixel in the initial pseudo-label can be assigned to any of the $N$ categories.

Second, since the extended prompt set introduces predictions beyond the target label space ($N_{in}$), we propose a refinement process that maps all non-target classes to the background class. Therefore, any pixel assigned to a class not included in the predefined set is re-labeled as background. 
The final pseudo-label $y^w_{out}$ is defined as:
\begin{equation}
\label{eq9}
y^w_{out}(i,j) =
\begin{cases}
C_n, & \text{if } C_n \leq C_{N_{in}} \\
\text{background}, & \text{otherwise}
\end{cases}
\end{equation}
where $C_n = y^w_{out, init}(i, j)$.
This process aligns the pseudo-labels with the label space required for the target semi-supervised segmentation task.
%
%
\begin{algorithm*}[ht!]
\caption{Pseudocode of SemiOVS (with FixMatch as the baseline)}
\label{algorithm}
\begin{algorithmic}[1]
\State \textbf{Input:} Labeled dataset $D^l$, Unlabeled In-Distribution (ID) dataset $D^{u}_{in}$, Unlabeled Out-of-Distribution (OOD) dataset $D^{u}_{out}$, Semi-Seg Model $f$, Open-Vocab Image Encoder $g_v$, Text Encoder $g_t$, Decoder $g_d$,Text prompt set $T$, Confidence thresholds $\tau_1, \tau_2$, Cross entropy $H$, Loss weight $\lambda$
\For{each mini-batch $(x^l, y^l) \sim D^l$, \; $x^u_{in} \sim D^{u}_{in}$, and $x^u_{out} \sim D^{u}_{out}$}
    \State \textcolor{softsky}{\# Supervised Flow}
    \State $p^l \gets f(x^l)$
    \State $L_s = H(p^l, y^l)$ \;\;\; (Eq. (\ref{eq2}))

    \State \textcolor{softsky}{{\# Unsupervised - In-Distribution Flow}}
    \State \textcolor{softgray}{\textit{Generate weak and strong perturbations for unlabeled ID data}}
    \State $x^w_{in}, \; x^s_{in} \gets \text{aug}_w(x^u_{in}), \; \text{aug}_s(x^u_{in})$

    \State $p^w_{in}, \; p^s_{in} \gets f(x^w_{in}), \; f(x^s_{in})$
    \State $y^w_{in} \gets \text{argmax}(p^w_{in})$
    \State $L_{u(in)} = \mathbbm{1}(max(p^w_{in}) \geq \tau_1) \cdot H(y^w_{in}, p^s_{in})$ \;\;\; (Eq. (\ref{eq3}))
    \\
    \State \textcolor{softsky}{\# Unsupervised - Out-of-Distribution Flow}
    \State \textcolor{softgray}{\textit{Generate weak and strong perturbations for unlabeled OOD data}}
    \State $x^w_{out}, \; x^s_{out} \gets \text{aug}_w(x^u_{out}), \; \text{aug}_s(x^u_{out})$
    \State \textcolor{softgray}{\textit{Compute the prediction on strong perturbation via the Semi-Seg Model}}
    \State $p^s_{out} \gets f(x^s_{out})$ \;\;\; (Eq. (\ref{eq6}))
    \\
    \State \textcolor{softgray}{\textit{Extract image embeddings (from weak perturbation) and text embeddings using OVS encoders}}
    \State $E_V \gets g_v(x^w_{out}) \in \mathbb{R}^{H \times W \times D}, \; E_T \gets g_t(T) \in \mathbb{R}^{N \times D}$ \;\;\; (Eq. (\ref{eq7}))
    \\
    \State \textcolor{softgray}{\textit{Generate pixel-level cost volume and corresponding embeddings from $E_V$ and $E_T$}}
    \State $C(i, j, n) \gets \frac{E_V(i,j) \cdot E_T(n)}{\lVert E_V(i,j) \rVert \; \lVert E_T(n) \rVert}$, \;\;$C_{emb}$ $\gets$ conv($C$) $\in \mathbb{R}^{H \times W \times N \times D}$  \;\;\; (Eq. (\ref{eq8}))
    \\
    \State \textcolor{softgray}{\textit{Calculate the prediction and the initial pseudo-label}}
    \State $p^w_{out}$ $\gets$ conv$(g_d(C_{emb}))$
    \State $y^w_{out, init} \gets \text{argmax}(p^w_{out})$
    \\
    \State \textcolor{softgray}{\textit{Refine the pseudo-label for OOD image using Eq. (\ref{eq9})}}
    \State $y^w_{out} \gets $ Eq. (\ref{eq9})($y^w_{out, init}$)
    \State $L_{u(out)} = \mathbbm{1}(max(p^w_{out}) \geq \tau_2) \cdot H(y^w_{out}, p^s_{out})$ \;\;\; (Eq. (\ref{eq4}))
    \\    
    \State \textcolor{softgray}{\textit{Update}} $f$ \textcolor{softgray}{\textit{with}} $L = L_s + L_{u(in)} + \lambda \cdot L_{u(out)}$ \;\;\; (Eq. (\ref{eq1}))
\EndFor
\end{algorithmic}
\end{algorithm*}

These strategies are based on the fact that OOD images may contain various objects beyond the predefined classes of the target task ($N_{in}$). For example, if an OOD image contains a giraffe, which is not among the predefined classes, using only the descriptions of classes within $N_{in}$ as the open-vocabulary text inputs could lead to misclassification.
The model might predict the giraffe as a horse or similar class within $N_{in}$, as it attempts to match the input to the closest available category.
This phenomenon can occur even when a background class is included in the text descriptions, as the predicted confidence score for classifying the giraffe as a background class might be lower than that for a horse or similar class.
By addressing such cases, we prevent class confusion and generate more accurate pseudo-labels, enabling the semi-supervised model to better classify ambiguous decision boundaries between in-distribution (i.e., predefined classes) and out-of-distribution classes (i.e., background).

In summary, using an OVS model for OOD images is crucial for two main reasons. First, it addresses the issue of class confusion arising from the presence of diverse objects in OOD images that are not part of the predefined classes. Second, it mitigates poor generalization performance caused by distribution discrepancies, even for objects that belong to the predefined classes. 
Consequently, by leveraging the ability of the OVS model to handle a broader range of semantic concepts, we can generate more reliable pseudo-labels for OOD images, enhancing the overall performance of the semi-supervised semantic segmentation task.
A pseudocode of our SemiOVS is presented in Algorithm \ref{algorithm}.

\subsection{More Details}
\label{sec:3.3}
To implement the proposed framework efficiently, we employ the SED \cite{xie2024sed} framework, the OVS method known for its rapid inference speed and competitive performance. Furthermore, we employ the multiple prompt templates strategy \cite{liang2023open,cho2024cat} to enhance the text representations. In particular, instead of using a single text description, multiple descriptions are used for each category, such as $\{$``a photo of a $\{C_n\}$, a bright photo of a $\{C_n\}$, ...''$\}$.
By applying this strategy, the dimensions of the text embeddings $E_T$ expand from $\mathbb{R}^{N \times D}$ to $\mathbb{R}^{N \times P \times D}$, and the initial cost volume increases from $\mathbb{R}^{H \times W \times N \times 1}$ to $\mathbb{R}^{H \times W \times N \times P}$, where $P$ is the number of prompt templates.

Moreover, we employ the class concept ensemble strategy proposed in the SemiVL \cite{hoyer2023semivl}. This strategy involves generating a set of concepts for each category, such as $\{$``dining table'', ``table for eating at''$\}$ for the ``dining table'' category, to provide rich semantic class information to the model. Therefore, we use class concept descriptions provided in SemiVL.
However, we use this approach slightly differently for more efficient inference. While SemiVL treats each class concept as a separate class and uses the class concept with the highest confidence score for aggregation, we average the concept embeddings at the text encoding step.
Formally, the class-concept text embedding $E_T \in \mathbb{R}^{N \times P \times D \times K}$ is averaged to a dimension of $\mathbb{R}^{N \times P \times D}$ as follows.
\begin{equation}
\label{eq10}
E_T = \frac{1}{K} \sum_{k=1}^C E_{T_k},
\end{equation}
where $K$ represents the number of concepts per class.

\section{Experiments}

\subsection{Datasets}
The datasets used in this study are categorized into in-distribution and OOD datasets. The in-distribution dataset, which is typically used for model training, includes labeled training, unlabeled training, and validation data, all of which share the same distribution characteristics. 
On the other hand, the OOD dataset consists exclusively of unlabeled training data and does not share the same distribution as the in-distribution data.

\subsubsection{In-Distribution Dataset}
The Pascal VOC 2012 dataset \cite{everingham2010pascal}, composed of images and corresponding object class labels, is widely used for semantic segmentation tasks. It contains 21 classes, including one background class, and the dataset is split into 1,464 images for training and 1,449 images for validation. Following previous studies~\cite{chen2021semi,wang2022semi,yang2023revisiting,zhao2023augmentation}, we used the augmented set~\cite{hariharan2011semantic}---with a total of 10,582 unique images used for training---in our experiment. The dataset was divided into two subsets based on the quality of annotations: the high-quality subset, consisting of 1,464 images with comprehensive annotations, and the coarse subset, containing 9,118 images with less detailed annotations. Additionally, three training protocols were considered for selecting labeled images, as outlined below:

\begin{itemize}
    \setlength{\itemindent}{-8pt}
    \item \textbf{Original:}
    This protocol uses labeled data from the high-quality subset, with the remaining data treated as unlabeled data. Experiments were conducted with 92, 183, 366, 732, and 1,464 labeled images. 
    \item \textbf{Blended:}
    This protocol involves the random selection of labeled data from the total dataset. A random subset of labeled data was selected at proportions of 1/16, 1/8, and 1/4 from the full dataset of 10,582 images, with the remaining images used as unlabeled data.
    \item \textbf{Priority:}
    This protocol prioritizes selecting labeled images from the high-quality subset. Labeled data was selected at proportions of 1/16, 1/8, and 1/4 from the full dataset of 10,582 images. If the high-quality subset is insufficient, additional images are selected from the coarse subset to complete the labeled data.
\end{itemize}

The Pascal Context \cite{mottaghi2014role} dataset provides additional annotations to the Pascal VOC 2010 dataset, making it more challenging for model training owing to its significantly larger number of object categories. Specifically, it includes over 400 object categories, in contrast to the Pascal VOC dataset. In this study, 59 object categories and one background category were selected, with the dataset consisting of 10,100 images. The dataset was split into 4,996 training images and 5,104 validation images. Since no predefined benchmark exists for splitting labeled and unlabeled data, the dataset was divided into labeled and unlabeled portions at proportions of 1/32, 1/16, 1/8, 1/4, and 1/2. For example, in the 1/32 split, 156 images were used as labeled data, while the remaining 4,840 images were used as unlabeled data.

\subsubsection{Out-of-Distribution Dataset}
\noindent\textbf{Web-scraped Images} 
To obtain additional unlabeled images, we used web sources containing a diverse range of images.
The web-scraping process was performed using the Selenium module of Python, which supports dynamic content scraping. Specifically, image crawling was performed using WebDriver of Selenium, which automated the process of entering queries into Google Images search page~\footnote{https://images.google.com/}. 
The search queries and the number of images collected for each class are provided in Appendix (Table~\ref{web_class_count}), and the scraped classes correspond to the same 20 categories within the Pascal VOC dataset. A total of 31,987 images were collected through web scraping but 62 images with excessively large resolutions were filtered out to ensure consistency and efficiency in data processing.

\noindent\textbf{Large-scale Dataset (COCO-Stuff)}
Another approach is to obtain additional unlabeled images from the publicly available COCO-Stuff dataset \cite{lin2014microsoft}. This dataset includes a wide variety of images, among which 20 classes align with the non-background categories in the Pascal VOC dataset, and 41 classes correspond to those in the Pascal Context dataset.
The dataset was divided into two settings: strict and filtered. In the strict setting, all images were used as additional unlabeled data. The filtered setting included only images containing the target classes, selected based on ground-truth information.

In conclusion, the COCO-Stuff dataset was used as additional unlabeled data in experiments with both the Pascal VOC and Pascal Context datasets. In the Pascal VOC experiments, 118,287 images were used in the strict setting, while 95,013 images were used in the filtered setting. Similarly, in the Pascal Context experiments, the strict setting employed 118,287 images, while 108,890 images were used in the filtered setting.

\begin{table*}[t!]
\caption{Performance evaluation on the \textit{Original} protocol of Pascal VOC. We evaluate the proposed method using three additional unlabeled OOD datasets. All methods use DeepLabV3+ with ResNet-50/101 backbone. The numbers in column headers (e.g., 92 and 183) indicate the quantity of labeled training images.}
\label{table:coco_web}
\centering
\resizebox{\linewidth}{!}{%
\begin{tabular}{lcc|ccccc}
\toprule
\multirow{2.5}{*}{\textbf{Pascal [\small{Original set}]}} & \multirow{2.5}{*}{Encoder} & \multirow{2.5}{*}{\parbox{3cm}{\centering Additional\\ Unlabeled Dataset}} & \multicolumn{5}{c}{\# Labeled images (Total: 10582)}\\ \cmidrule{4-8}
& & & 92 & 183 & 366 & 732 & 1464 \\
\midrule
UniMatch \cite{yang2023revisiting} & R-50 & - & 71.9 & 72.5 & 76.0 & 77.4 & 78.7 \\
\textbf{SemiOVS} (w/ UniMatch) & R-50 & COCO-Stuff-Strict (118k) & 77.7\small$\pm$0.2 \green{(\small{+5.8})} & 78.3\small$\pm$0.2 \green{(\small{+5.8})} & \textbf{79.6\small$\pm$0.1} \green{(\small{+3.6})} & 79.8\small$\pm$0.2 \green{(\small{+2.4})} & 79.4\small$\pm$0.2 \green{(\small{+0.7})} \\
\textbf{SemiOVS} (w/ UniMatch) & R-50 & COCO-Stuff-Filtered (95k) & \textbf{78.2\small$\pm$0.2} \green{(\small{+6.3})} & \textbf{78.7\small$\pm$0.1} \green{(\small{+6.2})} & 79.5\small$\pm$0.1 \green{(\small{+3.5})} & \textbf{80.0\small$\pm$0.2} \green{(\small{+2.6})} & \textbf{80.2\small$\pm$0.1} \green{(\small{+1.5})} \\
\textbf{SemiOVS} (w/ UniMatch) & R-50 & Web-scraped Data (32k) & 75.2\small$\pm$0.2 \green{(\small{+3.3})} & 78.5\small$\pm$0.1 \green{(\small{+6.0})} & 79.0\small$\pm$0.1 \green{(\small{+3.0})} & 79.4\small$\pm$0.1 \green{(\small{+2.0})} & 79.9\small$\pm$0.0 \green{(\small{+1.2})} \\
\midrule
UniMatch \cite{yang2023revisiting} & R-101 & - & 75.2 & 77.2 &	78.8 &	79.9 &	81.2 \\
\textbf{SemiOVS} (w/ UniMatch) & R-101 & COCO-Stuff-Strict (118k) & 80.1\small$\pm$0.2 \green{(\small{+4.9})} & 80.8\small$\pm$0.1 \green{(\small{+3.6})} & 81.5\small$\pm$0.2 \green{(\small{+2.7})} & \textbf{81.8\small$\pm$0.1} \green{(\small{+1.9})} & 81.5\small$\pm$0.1 \green{(\small{+0.3})} \\
\textbf{SemiOVS} (w/ UniMatch) & R-101 & COCO-Stuff-Filtered (95k) & \textbf{80.4\small$\pm$0.1} \green{(\small{+5.2})} & \textbf{81.3\small$\pm$0.2} \green{(\small{+4.1})} & \textbf{81.6\small$\pm$0.1} \green{(\small{+2.8})} & 81.7\small$\pm$0.1 \green{(\small{+1.8})} & \textbf{81.8\small$\pm$0.1} \green{(\small{+0.6})} \\
\textbf{SemiOVS} (w/ UniMatch) & R-101 & Web-scraped Data (32k) & 77.4\small$\pm$0.2 \green{(\small{+2.2})} & 80.3\small$\pm$0.2 \green{(\small{+3.1})} & 80.4\small$\pm$0.2 \green{(\small{+1.6})} & 80.8\small$\pm$0.1 \green{(\small{+0.9})} & 81.3\small$\pm$0.0 \green{(\small{+0.1})} \\
\bottomrule
\end{tabular}
}
\end{table*}

\begin{table*}[th!]
\caption{Comparison with state-of-the-art methods on the Pascal VOC \textit{Original} protocol. All methods (except those marked with $\dagger$) are trained using the ResNet-50/101 encoder and the DeepLabV3+ decoder. Bold and underline indicate the best and second-best results, respectively.}
\label{table:original}
\centering
\resizebox{0.95\linewidth}{!}{%
\begin{tabular}{lc|ccccc}
\toprule
\multirow{2.5}{*}{\textbf{Pascal [\small{Original set}]}} & \multirow{2.5}{*}{Encoder} & \multicolumn{5}{c}{\# Labeled images (Total: 10582)}\\ \cmidrule{3-7}
& & 92 & 183 & 366 & 732 & 1464 \\
\midrule
Supervised Baseline & R-50 & 44.0 & 52.3 & 61.7 & 66.7 & 72.9 \\
PseudoSeg \cite{zou2020pseudoseg} & R-50 & 54.9 & 61.9 &	64.9 &	70.4 & 71.0 \\
PC$^{2}$Seg \cite{zhong2021pixel} & R-50 & 56.9 & 64.6 &	67.6 &	70.9 & 72.3 \\
CPCL \cite{fan2023conservative} & R-50 & 61.9 & 67.0 & 72.1 & 74.3 & - \\
AugSeg \cite{zhao2023augmentation} & R-50 & 64.2 & 72.2 & 76.2 &	77.4 & 78.8 \\
NP-SemiSeg \cite{wang2023np} & R-50 & 65.8 & 72.4 & 75.8 & 77.4 & - \\
Dual Teacher \cite{na2023switching} & R-50 & 70.8 & 74.5 & 76.4 & 77.7 & 78.2 \\
\midrule
FixMatch \cite{sohn2020fixmatch,yang2023revisiting} & R-50 & 63.8 & 70.3 & 73.2 & 76.8 & 77.8\\
\textbf{SemiOVS} (w/ FixMatch) & R-50 & \underline{76.4\small$\pm$0.4} \green{(\small{+12.6})} & 77.8\small$\pm$0.3 \green{(\small{+7.5})} & 78.9\small$\pm$0.1 \green{(\small{+5.7})} & 79.1\small$\pm$0.2 \green{(\small{+2.3})} & \underline{79.5\small$\pm$0.3} \green{(\small{+1.7})} \\
\midrule
UniMatch \cite{yang2023revisiting} & R-50 & 71.9 & 72.5 & 76.0 &	77.4 & 78.7 \\
\textbf{SemiOVS} (w/ UniMatch) & R-50 & \textbf{78.2\small$\pm$0.2} \green{(\small{+6.3})} & \underline{78.7\small$\pm$0.1} \green{(\small{+6.2})} & \underline{79.5\small$\pm$0.1} \green{(\small{+3.5})} & \underline{80.0\small$\pm$0.2} \green{(\small{+2.6})} & \textbf{80.2\small$\pm$0.1} \green{(\small{+1.5})} \\
\midrule
PrevMatch \cite{shin2024revisiting} & R-50 & 73.4 & 75.4 & 77.5 & 78.6 & 79.3 \\
\textbf{SemiOVS} (w/ PrevMatch) & R-50 & \textbf{78.2\small$\pm$0.2} \green{(\small{+4.8})} & \textbf{79.1\small$\pm$0.2} \green{(\small{+3.7})} & \textbf{79.6\small$\pm$0.1} \green{(\small{+2.1})} & \textbf{80.1\small$\pm$0.2} \green{(\small{+1.5})} & \textbf{80.2\small$\pm$0.1} \green{(\small{+0.9})} \\
\midrule
\midrule
Supervised Baseline	& R-101 & 45.1 &	55.3 &	64.8 &	69.7 &	73.5 \\
CPS \cite{chen2021semi} & R-101 & 64.1 & 67.4 & 71.7 & 75.9 &	- \\
ReCo \cite{liu2021bootstrapping} & R-101 & 64.8 & 72.0 & 73.1 & 74.7 & - \\
U$^{2}$PL \cite{wang2022semi}	& R-101 &	68.0 & 69.2 & 73.7 & 76.2 &	79.5 \\
GTA-Seg \cite{jin2022semi}	& R-101 &	70.0 &	73.2 &	75.6 &	78.4 & 80.5 \\
PCR	\cite{xu2022semi}	& R-101 &	70.1 & 74.7 & 77.2 & 78.5 &	80.7 \\
DGCL \cite{wang2023hunting} & R-101 &	70.5 & 77.1 & 78.7 & 79.2 & 81.6 \\
CCVC \cite{wang2023conflict}	& R-101 & 70.2 & 74.4 & 77.4 & 79.1 &	80.5 \\
LogicDiag \cite{liang2023logic} & R-101 & 73.3 &	76.7 & 77.9 & 79.4 & - \\
Diverse Co-T. (3-cps) \cite{li2023diverse}	& R-101 & 75.7 & 77.7 & 80.1 & 80.9 & 82.0 \\
PRCL \cite{xie2024prcl} & R-101 & 70.2 & 72.2 & 75.2 & 76.2 & 78.3 \\
U$^{2}$PL+ \cite{wang2024using} & R-101 & 69.3 & 73.4 & 75.0 & 77.1 & 79.5 \\
CorrMatch \cite{sun2024corrmatch} & R-101 & 76.4 & 78.5 & 79.4 & 80.6 & 81.8 \\
DDFP \cite{wang2024towards} & R-101 & 75.0 & 78.0 & 79.5 & 81.2 & 82.0 \\
AllSpark$^\dagger$ \cite{wang2024allspark} & MiT-B5 & 76.1 & 78.4 & 79.8 & 80.8 & \underline{82.1} \\
\midrule
UniMatch \cite{yang2023revisiting}	& R-101 & 75.2 & 77.2 &	78.8 &	79.9 &	81.2 \\
\textbf{SemiOVS} (w/ UniMatch) & R-101 & 80.4\small$\pm$0.1 \green{(\small{+5.2})} & 81.3\small$\pm$0.2 \green{(\small{+4.1})} & \underline{81.6\small$\pm$0.1} \green{(\small{+2.8})} & 81.7\small$\pm$0.1 \green{(\small{+1.8})} & 81.8\small$\pm$0.1 \green{(\small{+0.6})} \\
\midrule
PrevMatch \cite{shin2024revisiting} & R-101 & 77.0 & 78.5 & 79.6 & 80.4 & 81.6 \\
\textbf{SemiOVS} (w/ PrevMatch) & R-101 & \underline{80.5\small$\pm$0.1} \green{(\small{+3.5})} & \underline{81.5\small$\pm$0.1} \green{(\small{+3.0})} & 81.4\small$\pm$0.2 \green{(\small{+1.8})} & \underline{81.8\small$\pm$0.2} \green{(\small{+1.4})} & 81.9\small$\pm$0.1 \green{(\small{+0.3})} \\
\midrule
SemiVL$^\dagger$ \cite{hoyer2023semivl} & ViT-B/16 & 84.0 & 85.6 & 86.0 & 86.7 & 87.3 \\
\textbf{SemiOVS} (w/ SemiVL$^\dagger$) & ViT-B/16 & \textbf{87.0\small$\pm$0.1} \green{(\small{+3.0})} & \textbf{87.3\small$\pm$0.2} \green{(\small{+1.7})} & \textbf{87.5\small$\pm$0.2} \green{(\small{+1.5})} & \textbf{87.9\small$\pm$0.1} \green{(\small{+1.2})} & \textbf{88.0\small$\pm$0.1} \green{(\small{+0.7})} \\
\bottomrule
\end{tabular}
}
\end{table*}

\begin{table}[ht!]
\caption{Comparison with state-of-the-art methods on the Pascal VOC \textit{Blended} protocol. The fractional values in the column headers represent the ratio of labeled training images.}
\label{table:blended}
\centering
\resizebox{\linewidth}{!}{%
\begin{tabular}{lc|ccc}
\toprule
\textbf{Pascal [\small{Blended set}}] & Encoder & 1/16 & 1/8 & 1/4\\
\midrule
Supervised Baseline & R-50 & 62.4 & 68.2 & 72.3\\
Mean Teacher \cite{tarvainen2017mean} & R-50 & 66.8 & 70.8 & 73.2\\
CPS \cite{chen2021semi} & R-50 &72.0 & 73.7 & 74.9 \\
PS-MT \cite{liu2022perturbed} & R-50 & 72.8 & 75.7 & 76.4 \\
CCVC \cite{wang2023conflict} & R-50 & 74.5 & 76.1 & 76.4 \\
iMAS \cite{zhao2023instance} & R-50 & 74.8 & 76.5 & 77.0 \\
GPS \cite{cho2024interactive} & R-50 & 72.9 & 75.7 & 76.3 \\
RWMS \cite{liu2024rwms} & R-50 & 72.2 & 74.9 & 75.1 \\
UniMatch \cite{yang2023revisiting} & R-50 & 75.8 & 76.9 & 76.8 \\
\midrule
PrevMatch \cite{shin2024revisiting} & R-50 & \underline{76.0} & \underline{77.1} & \underline{77.6} \\
\multirow{2}{*}{\textbf{SemiOVS} (w/ PrevMatch)} & \multirow{2}{*}{R-50} & \textbf{78.9} & \textbf{79.2} & \textbf{78.9}\\
& & \green{(\small{+2.9})} & \green{(\small{+2.1})} & \green{(\small{+1.3})}\\
\midrule
\midrule
Supervised Baseline	& R-101 & 67.5 & 71.1 & 74.2 \\
CPS \cite{chen2021semi} & R-101 & 74.5 & 76.4 & 77.7 \\
U$^2$PL \cite{wang2022semi} & R-101 & 74.4 & 77.6 & 78.7 \\
CISC-R \cite{wu2023querying} & R-101 & 75.3 & 77.1 & 77.2 \\
Diverse Co-T. (3-cps) \cite{li2023diverse} & R-101 & 77.6 & 79.0 & 80.0 \\
CorrMatch \cite{sun2024corrmatch} & R-101 & \underline{78.4} & 79.3 & 79.6 \\
DDFP \cite{wang2024towards} & R-101 & 78.3 & 78.9 & 79.8 \\
AllSpark \cite{wang2024allspark} & MiT-B5 & 78.3 & \underline{80.0} & \underline{80.4} \\
\midrule
UniMatch \cite{yang2023revisiting} & R-101 & 78.1 & 78.4 & 79.2  \\
\multirow{2}{*}{\textbf{SemiOVS} (w/ UniMatch)} & \multirow{2}{*}{R-101} & \textbf{80.3} & \textbf{80.4} & \textbf{80.7}\\
& & \green{(\small{+2.2})} & \green{(\small{+2.0})} & \green{(\small{+1.5})}\\
\midrule
\bottomrule
\end{tabular}
}
\end{table}

\begin{table}[ht!]
\caption{Comparison with state-of-the-art methods on the Pascal VOC \textit{Priority} protocol.}
\label{table:prioritizing}
\centering
\resizebox{\linewidth}{!}{%
\begin{tabular}{lc|ccc}
\toprule
\textbf{Pascal [\small{Priority set}}] & Encoder & 1/16 & 1/8 & 1/4\\
\midrule
Supervised Baseline & R-101 & 70.6 & 75.0 & 76.5 \\
U$^{2}$PL \cite{wang2022semi} & R-101 & 77.2 & 79.0 & 79.3 \\
AugSeg \cite{zhao2023augmentation} & R-101 & 79.3 & 81.5 & 80.5 \\
Dual Teacher \cite{na2023switching} & R-101 & 80.1 & 81.5 & 80.5 \\
CorrMatch \cite{sun2024corrmatch} & R-101 & 81.3 & 81.9 & 80.9\\
AllSpark \cite{wang2024allspark} & MiT-B5 & \underline{81.6} & 82.0 & 80.9\\
\midrule
UniMatch \cite{yang2023revisiting} & R-101 & 80.9 & 81.9 & 80.4 \\
\multirow{2}{*}{\textbf{SemiOVS} (w/ UniMatch)} & \multirow{2}{*}{R-101} & 81.5 & \textbf{82.4} & \underline{81.0}\\
& & \green{(\small{+0.6})} & \green{(\small{+0.5})} & \green{(\small{+0.6})}\\
\midrule
PrevMatch \cite{shin2024revisiting} & R-101 & 81.4 & 81.9 & 80.8\\
\multirow{2}{*}{\textbf{SemiOVS} (w/ PrevMatch)} & \multirow{2}{*}{R-101} & \textbf{82.0} & \underline{82.3} & \textbf{81.3}\\
& & \green{(\small{+0.6})} & \green{(\small{+0.4})} & \green{(\small{+0.5})}\\
\midrule
\bottomrule
\end{tabular}
}
\end{table}

\subsection{Implementation Details}
We selected FixMatch \cite{sohn2020fixmatch}, UniMatch \cite{yang2023revisiting}, PrevMatch \cite{shin2024revisiting}, and SemiVL \cite{hoyer2023semivl} as baseline models to comprehensively evaluate the proposed method. For FixMatch, UniMatch, and PrevMatch, ResNet-50 or ResNet-101 \cite{he2016deep} was used as the encoder, with a DeepLabV3+ \cite{chen2018encoder} decoder. These models were trained under a consistent setup, with each mini-batch comprising 8 labeled images, 8 in-distribution unlabeled images, and 8 OOD unlabeled images. Training was conducted for 80 epochs on both the Pascal VOC and Pascal Context datasets using the SGD optimizer on a single GPU. The initial learning rate was set to 0.001 and adjusted during training with a polynomial learning rate scheduler. Random cropping was applied, with crop sizes of 321×321 or 513×513 pixels for the Pascal VOC dataset and 321×321 pixels for the Pascal Context dataset. 
For SemiVL, ViT-B/16 was employed as the encoder, along with a language-guided decoder that incorporates additional text inputs. The training setup for SemiVL used mini-batches comprising 4 labeled images, 4 in-distribution unlabeled images, and 4 OOD unlabeled images. Training was performed for 80 epochs on the Pascal VOC dataset using the AdamW optimizer with an initial learning rate of 0.0001 on two GPUs. Random cropping was applied with a crop size of 512×512.

Across all baselines, we used common weak(e.g., resize, crop, and flip) and strong(e.g., color transformations, grayscale, CutMix, and blur) data augmentations, following UniMatch~\cite{yang2023revisiting}. The weight $\lambda$ for the OOD flow was set to 1.0 for FixMatch, UniMatch, and PrevMatch, and 0.5 for SemiVL. Most of the experimental results reported in this paper are averaged over two or three runs to ensure reliability.
For the confidence threshold, $\tau$ was set to 0.95 for in-distribution flow and 0 for OOD flow. 
For the OVS process, we employed an offline semi-supervised learning approach, where pseudo-labels are pre-generated, to enhance training efficiency. 
For the number of candidate classes ($N$) used in the text prompt set of the OVS process, we selected 171 candidate classes from the COCO-Stuff dataset~\footnote{https:/github.com/nightrome/cocostuff/blob/master/labels.md} for Pascal VOC and 162 candidate classes for Pascal Context. In the case of Pascal Context, certain categories, such as ``wall'', are represented as a single class, whereas COCO-Stuff distinguishes them into multiple material-based subcategories (e.g., wall-brick, wall-concrete, wall-wood). Therefore, we manually mapped these subcategories to a single category in Pascal Context. These text prompts will be released together with the source code.
Experiments were conducted on Ubuntu 20.04, Python 3.9, PyTorch 1.13, CUDA 11.3, and NVIDIA RTX 3090Ti or A6000 GPUs.

The evaluation metric used in this study was mIoU. The IoU is calculated for each class across all images and then averaged to obtain the mIoU score, which represents the performance of the model across all classes. 
The formula for mIoU is provided in Eq.~(\ref{mIoU}), where $n$ is the total number of classes.
\begin{align} 
  \begin{array} {cl}    
    \text{IoU} = \frac{\text{Area of Intersection}}{\text{Area of Union}} \\ \\
    \text{mIoU} = \frac{\text{IoU}_1 + \text{IoU}_2 + \dots + \text{IoU}_n}{n}
  \end{array} 
  \label{mIoU}
\end{align}

\subsection{Results}
We conducted experiments on three different additional unlabeled OOD datasets based on the UniMatch baseline. 
As shown in Table \ref{table:coco_web}, the proposed method achieves significant performance improvements across all protocols compared to the baseline. In particular, in the COCO-Filtered setting, we observed substantial improvements of 6.3\% and 6.2\% in the 92 and 183 settings (R-50), respectively. Of the three additional OOD datasets, COCO-Filtered shows the best overall performance gains. While using web-scraped images also improves the performance, the gains are relatively smaller than those achieved with COCO images. This suggests that the pseudo-labels for web-scraped images may be relatively less accurate, possibly owing to greater diversity and noise in these images.

\begin{table*}[ht!]
\caption{Evaluation results on the Pascal Context dataset.}
\label{table:pc59}
\centering
\resizebox{\linewidth}{!}{%
\begin{tabular}{lc|ccccc}
\toprule
\multirow{2.5}{*}{\textbf{Pascal Context}} & \multirow{2.5}{*}{Encoder} & 1/32 & 1/16 & 1/8 & 1/4 & 1/2 \\
& & (156) & (312) & (624) & (1249) & (2498) \\
\midrule
Supervised Baseline & R-50 & 26.4 & 30.0 & 34.4 & 38.0 & 40.9 \\
\midrule
FixMatch \cite{sohn2020fixmatch,yang2023revisiting} & R-50 & 29.6 & 35.1 & 37.9 & 40.3 & 42.0 \\
\textbf{SemiOVS} (w/ FixMatch) & R-50 & 40.8\small$\pm$0.5 \green{(\small{+11.2})} & 42.0\small$\pm$0.3 \green{(\small{+6.9})} & 43.2\small$\pm$0.3 \green{(\small{+5.3})} & 44.0\small$\pm$0.2 \green{(\small{+3.7})} & 44.1\small$\pm$0.1 \green{(\small{+2.1})} \\
\midrule
UniMatch \cite{yang2023revisiting} & R-50 & 33.4 & 37.0 & 39.2 & 41.2 & 42.3 \\
\textbf{SemiOVS} (w/ UniMatch) & R-50 & \textbf{42.3\small$\pm$0.2} \green{(\small{+8.9})} & \underline{43.6\small$\pm$0.4} \green{(\small{+6.6})} & \underline{44.3\small$\pm$0.2} \green{(\small{+5.1})} & \underline{45.1\small$\pm$0.3} \green{(\small{+3.9})} & \textbf{45.2\small$\pm$0.1} \green{(\small{+2.9})} \\
\midrule
PrevMatch \cite{shin2024revisiting} & R-50 & 33.5 & 37.3 & 39.2 & 41.4 & 42.1 \\
\textbf{SemiOVS} (w/ PrevMatch) & R-50 & \underline{42.1\small$\pm$0.3} \green{(\small{+8.6})} & \textbf{43.8\small$\pm$0.3} \green{(\small{+6.5})} & \textbf{44.5\small$\pm$0.2} \green{(\small{+5.3})} & \textbf{45.2\small$\pm$0.1} \green{(\small{+3.8})} & \underline{44.8\small$\pm$0.2} \green{(\small{+2.7})} \\
\midrule
\midrule
Supervised Baseline	& R-101 & 28.1 & 32.5 & 36.2 & 39.7 & 42.9 \\
\midrule
UniMatch \cite{yang2023revisiting} & R-101 & 36.0 & 39.6 & 41.5 & 43.1 & 43.7 \\
\textbf{SemiOVS} (w/ UniMatch) & R-101 & \underline{42.6\small$\pm$0.5} \green{(\small{+6.6})} & \underline{45.1\small$\pm$1.1} \green{(\small{+5.5})} & \underline{46.3\small$\pm$0.3} \green{(\small{+4.8})} & \underline{47.3\small$\pm$0.3} \green{(\small{+4.2})} & \underline{47.3\small$\pm$0.2} \green{(\small{+3.6})} \\
\midrule
PrevMatch \cite{shin2024revisiting} & R-101 & 35.7 & 39.9 & 41.6 & 43.3 & 44.0\\
\textbf{SemiOVS} (w/ PrevMatch) & R-101 & \textbf{44.1\small$\pm$0.6} \green{(\small{+8.4})} & \textbf{46.2\small$\pm$0.5} \green{(\small{+6.3})} & \textbf{46.5\small$\pm$0.3} \green{(\small{+4.9})} & \textbf{47.9\small$\pm$1.0} \green{(\small{+4.6})} & \textbf{48.0\small$\pm$0.2} \green{(\small{+4.0})} \\
\bottomrule
\end{tabular}
}
\end{table*}

\subsection{Comparison with State-of-the-Art Methods}
Unless otherwise specified, subsequent experiments were conducted primarily in the COCO-Stuff-Filtered setting.

Table \ref{table:original} lists the results, compared with state-of-the-art methods on the \textit{Original} protocol of Pascal VOC.
To comprehensively validate the proposed method, we adopted various methods (FixMatch, UniMatch, PrevMatch, and SemiVL) as baselines. First, applying our SemiOVS method to each baseline significantly improves performance across all methods. Notably, SemiOVS with FixMatch (R-50) shows a substantial improvement of 12.6\% over the baseline (FixMatch) in the 92-label setting. 
Moreover, our method outperforms existing state-of-the-art methods in the R-101 and 92-label settings by more than 4\%. Even compared to SemiVL, which exhibits strong performance through vision-language-based models, SemiOVS achieves a 3\% performance gain in the 92-label setting. 
Furthermore, Tables \ref{table:blended} and \ref{table:prioritizing} present results on the \textit{Blended} and \textit{Priority} protocols, respectively. The proposed method significantly improves performance, achieving state-of-the-art results in these protocols.

To validate the proposed method in a more complex context, we performed experiments on the Pascal Context dataset.
Since there are no predefined benchmark protocols for semi-supervised segmentation, we reproduced the results of several baseline methods (FixMatch, UniMatch, and PrevMatch) and reported the results by applying SemiOVS to each baseline, as listed in Table \ref{table:pc59}.
Similar to the results on Pascal VOC, the proposed SemiOVS achieves substantial improvements over the baseline methods, with gains ranging from 6.6 to 10.2\% in the 1/32 protocol and 2.1 to 4.0\% in the 1/2 protocol.
These results suggest that using additional unlabeled OOD images and the OVS model aids in capturing class boundaries in more complex contexts.

\section{Analysis}
\subsection{Ablation studies of the proposed method}
We conducted ablation studies (Table \ref{table:replace_ovs_with_sss}), to verify the effectiveness of our two main contributions: (1) using additional unlabeled images and (2) incorporating an OVS model into the semi-supervised segmentation framework.
The first row indicates the baseline (UniMatch). First, to investigate the efficacy of using additional unlabeled images, we replaced the OVS model with a semi-supervised segmentation model. In other words, we fed both in-distribution unlabeled images and additional unlabeled OOD images into the same semi-supervised segmentation model.
The results exhibit significant improvements in the 92 and 183 label settings, but performance degradation in the 732 and 1464 label settings. These results suggest that in scenarios with fewer labels, where models may struggle to learn class boundaries due to label scarcity, the diverse visual patterns in OOD images can enhance the learning ability of the model. On the other hand, in scenarios with relatively abundant labels (732 and 1464), where the model already performs well on the main classification task, the distribution gap between in-distribution and OOD images causes noise in pseudo-labels, which can hinder the learning process and lead to performance degradation.

The third row presents the results of our proposed framework (i.e., SemiOVS), which leverages the OVS model to process OOD images. The results show even larger performance gains in scenarios with fewer labels (92 and 183) while also achieving significant improvements in scenarios with relatively abundant labels (732 and 1464).
This suggests that by effectively addressing two potential challenges from OOD images---class confusion from diverse out-of-class objects and distribution discrepancies of in-class objects---the OVS model generates more accurate pseudo-labels. 
Consequently, this process enables the semi-supervised segmentation model to learn stably from more accurate and abundant supervision signals.
\begin{table}[t]
\caption{Ablation study of the proposed method using the UniMatch method. The scores below the teacher (e.g., 79.0 and 78.9) represent the teacher's IoU scores on the Pascal VOC validation set.}
\label{table:replace_ovs_with_sss}
\centering
\resizebox{\linewidth}{!}{%
\begin{tabular}{cc|ccccc}
\toprule
Teacher & Student & 92 & 183 & 366 & 732 & 1464\\
\midrule
- & R-101, DLv3+ & 75.2 & 77.2 & 78.8 & 79.9 & 81.2 \\
\midrule
Semi-Seg Model & \multirow{2}{*}{R-101, DLv3+} & 78.0 & 78.2 & 78.9 & 79.2 & 79.2\\
(79.0) & & \green{(\small{+2.8})} & \green{(\small{+1.0})} & \green{(\small{+0.1})} & \red{(\small{-0.7})} & \red{(\small{-2.0})}\\
\midrule
OVS Model & \multirow{2}{*}{R-101, DLv3+} & 80.4 & 81.3 & 81.6 & 81.7 & 81.8 \\
(78.9) & & \green{(\small{+5.2})} & \green{(\small{+4.1})} & \green{(\small{+2.8})} & \green{(\small{+1.8})} & \green{(\small{+0.6})}\\
\bottomrule
\end{tabular}
}
\end{table}

\subsection{Effect of the class concept ensemble}
As described in Section \ref{sec:3.3}, we utilized multiple concepts for each category as text descriptions for the text encoder of the OVS model. Table \ref{table:ablation_concept_ensemble} shows that while using only class names significantly improves performance over the baseline, leveraging the concept ensemble further enhances it.
These results suggest that using multiple concepts to describe a single class helps the model better understand category characteristics by providing diverse semantic descriptions, rather than relying solely on a single class name.
\begin{table}[h]
\caption{Ablation study on the efficacy of the class concept ensemble using the UniMatch method and ResNet-50 encoder.}
\label{table:ablation_concept_ensemble}
\centering
\resizebox{\linewidth}{!}{%
\begin{tabular}{l|ccccc}
\toprule
Class Definition & 92 & 183 & 366 & 732 & 1464\\
\midrule
Baseline & 71.9 & 72.5 & 76.0 &	77.4 & 78.7 \\
\midrule
Class Name (Single) & 77.1 & 77.8 & 78.9 & 79.5 & 80.0 \\
Concepts Ensemble & 78.2 & 78.7 & 79.5 & 80.0 & 80.2 \\
\bottomrule
\end{tabular}
}
\end{table}

\subsection{Effect of the number of unlabeled OOD images}
We performed experiments to investigate the effect of the quantity of additional unlabeled OOD images.
The results in Table \ref{table:amount_of_images} show that using more unlabeled images generally leads to better performance. Although using all unlabeled images results in the best performance, we observe substantial improvements over the baseline even with a relatively small number of additional unlabeled images, such as COCO (10k) or Web (16k).
This indicates that the proposed method can achieve significant performance gains without requiring an extensive amount of additional unlabeled data, demonstrating its practical applicability in real-world scenarios.
%
\begin{table}[t]
\caption{Ablation study on the number of additional unlabeled OOD images using the UniMatch method and ResNet-50 encoder.}
\label{table:amount_of_images}
\centering
\resizebox{\linewidth}{!}{%
\begin{tabular}{c|ccccc}
\toprule
Number of Unlabeled Images & 92 & 183 & 366 & 732 & 1464\\
\midrule
Baseline & 71.9 & 72.5 & 76.0 &	77.4 & 78.7 \\
\midrule
COCO (10k) & 77.1 & 78.4 & 78.5 & 79.2 & 79.4 \\
COCO (30k) & 77.4 & 78.7 & 79.2 & 79.7 & 79.8 \\
COCO (50k) & 77.7 & 78.7 & 79.5 & 79.6 & 79.9 \\
COCO (95k) & 78.2 & 78.7 & 79.5 & 80.0 & 80.2 \\
Web (16k) & 74.6 & 78.6 & 78.8 & 79.3 & 79.4\\
Web (32k) & 75.2 & 78.5 & 79.0 & 79.4 & 79.9 \\
\bottomrule
\end{tabular}
}
\end{table}

\begin{table}[b]
\caption{Comparison of performance and model parameters between the OVS model and the proposed method.}
\label{table:comparision_params}
\centering
\resizebox{\linewidth}{!}{%
\begin{tabular}{ccc|ccc}
\toprule
Method & Encoder & \#Params & 92 & \;366 & 1464\\
\midrule
OVS Model & ConvNeXt-L & 353.2M & \multicolumn{3}{c}{Val IoU: 78.9} \\  
\midrule
SemiOVS (w/ UniM.) & R-50 & 40.5M & 78.2 & 79.5 & 80.2 \\
SemiOVS (w/ PrevM.) & R-101 & 59.5M & 80.5 & 81.4 & 81.9 \\
SemiOVS (w/ SemiVL) & ViT-B/16 & 89.0M & 87.0 & 87.5 & 88.0 \\
\bottomrule
\end{tabular}
}
\end{table}
\subsection{Performance and parameter comparison with OVS model}
We compared the performance and efficiency between the OVS and our semi-supervised segmentation models.
OVS models generally demonstrate strong generalization performance across different datasets; however, they require large-scale models to maintain their dataset-agnostic capabilities. As shown in Table \ref{table:comparision_params}, while the pretrained OVS model achieves a good zero-shot performance of 78.9\% on the Pascal VOC validation set, it requires 353.2M parameters. In contrast, the proposed method surpasses the OVS model with only 40.5M parameters (approximately 9 times fewer). Furthermore, SemiOVS (w/ SemiVL) outperforms the OVS model by a significant margin of approximately 9\% while requiring approximately 4 times fewer parameters. Although these models may not generalize across various datasets like the OVS model, they effectively perform their target task of segmenting predefined classes within a specific dataset.

\begin{table}[h]
\caption{Ablation study on the confidence threshold $\tau$ for pseudo-labeling OOD images (Eq. (\ref{eq4})), using the UniMatch method with a ResNet-50 encoder. $\dagger$ denotes the default value ($\tau$ = 0.0) used in the main experiments.}
\label{table:confidence_threshold}
\centering
\resizebox{\linewidth}{!}{%
\begin{tabular}{c|ccccc}
\toprule
Confidence Threshold ($\tau$) & 92 & 183 & 366 & 732 & 1464\\
\midrule
Baseline & 71.9 & 72.5 & 76.0 &	77.4 & 78.7 \\
\midrule
0.0$^\dagger$ & 78.2 & 78.7 & 79.5 & \underline{80.0} & \underline{80.2} \\
0.1 & \textbf{78.5} & \underline{78.9} & \textbf{79.8} & \textbf{80.1} & \underline{80.2} \\
0.25 & \underline{78.4} & \textbf{79.3} & \underline{79.6} & \textbf{80.1} & \textbf{80.3} \\
0.5 & 77.7 & 78.7 & 79.5 & \textbf{80.1} & \underline{80.2} \\ 
0.75 & 77.0 & 78.1 & 78.9 & 79.7 & 79.8 \\ 
0.9 & 75.7 & 77.5 & 78.6 & 79.3 & 79.4 \\
0.95 & 74.7 & 77.5 & 78.3 & 78.7 & 79.1 \\
\bottomrule
\end{tabular}
}
\end{table}
\subsection{Effect of the confidence threshold}
We conducted experiments across a wide range of values from 0.0 to 0.95, as presented in Table \ref{table:confidence_threshold}, to investigate the impact of the confidence threshold ($\tau$) used in Eq. (\ref{eq4}).  All threshold settings outperform the baseline, indicating the effectiveness of incorporating the OOD flow. Notably, thresholds between 0.0 (not filtering) and 0.5 consistently yield strong performance across various label settings. However, when $\tau$ exceeds 0.5, the performance gain diminishes, particularly under low-label scenarios (e.g., 92 or 183 labels).
This phenomenon is attributed to the trade-off between label precision and data diversity: as $\tau$ increases, only high-confidence (and thus more accurate) pseudo-labels are retained, improving label quality. However, a large portion of the OOD regions is excluded, reducing the amount of diverse visual information available for training. Consequently, the model may miss out on learning from a broader range of object appearances and contexts that are beneficial for generalization.
Although we adopted $\tau$=0.0 for simplicity in our main experiments, these results suggest that tuning $\tau$ within the range of [0.0, 0.5] could further improve performance.

\begin{figure*}[t!]
\centering
\includegraphics[width=1.0\linewidth]{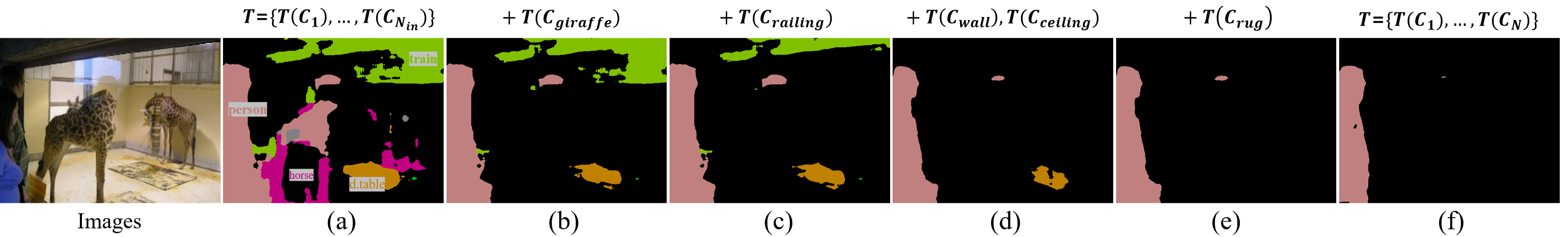}
\caption{Visualization of pseudo-labels generated based on different text prompt sets ($T$) in the OVS model. (a) represents the case where only in-distribution classes from Pascal VOC, including the background class, were used. (b)--(e) show the progressive addition of candidate classes: giraffe, railing, wall and ceiling, and rug, respectively. (f) corresponds to the final setting used in this study.}
\label{fig:ovs_step_by_step}
\end{figure*}

\subsection{Effect of loss weight}
We investigated the effect of the loss weight $\lambda$ in Eq. (\ref{eq5}), which controls the contribution of the OOD flow during training. The results, summarized in Table \ref{table:loss_weight}, show consistent improvements over the baseline across a broad range of $\lambda$ values from 0.1 to 5.0, demonstrating the effectiveness of the proposed flow.
In particular, values in the range of 0.5 to 2.0 achieve the best overall performance. By contrast, setting $\lambda$ too low (e.g., 0.1) fails to sufficiently leverage the additional supervision from pseudo-labeled OOD images, resulting in limited gains. Conversely, excessively large values (e.g., 5.0) cause the model to overly rely on OOD signals, which may lead to overfitting to visual patterns that are less relevant to the target domain (i.e., Pascal VOC), ultimately degrading task-specific performance. These results suggest that moderate values of $\lambda$ (e.g., 0.5–2.0) provide a robust balance between generalization ability and task-specific performance.
\begin{table}[t]
\caption{Ablation study on the loss weight $\lambda$ for the OOD flow (Eq. (\ref{eq5})), using the UniMatch method with a ResNet-50 encoder. $\dagger$ denotes the default value ($\lambda$ = 1.0) used in the main experiments.}
\label{table:loss_weight}
\centering
\resizebox{\linewidth}{!}{%
\begin{tabular}{c|ccccc}
\toprule
Loss Weight ($\lambda$) & 92 & 183 & 366 & 732 & 1464\\
\midrule
Baseline & 71.9 & 72.5 & 76.0 &	77.4 & 78.7 \\
\midrule
0.1 & 75.6 & 76.8 & 78.3 & 79.6 & \underline{80.1} \\
0.5 & \underline{77.6} & 78.2 & \textbf{79.6} & \underline{79.8} & \textbf{80.2} \\
1.0$^\dagger$ & \textbf{78.2} & \underline{78.7} & \underline{79.5} & \textbf{80.0} & \textbf{80.2} \\
2.0 & 77.5 & \textbf{78.9} & 79.2 & 79.7 & 79.6 \\
5.0 & 75.9 & 77.2 & 77.2 & 78.1 & 77.4 \\
\bottomrule
\end{tabular}
}
\end{table}

\subsection{Effect of text prompt set construction}
\label{subsec:prompt_set}
An appropriate class prompt set must be used (as discussed regarding Eq. \ref{eq9} in Section \ref{subsec:3.2}) to generate effective pseudo-labels for semi-supervised semantic segmentation using the OVS model. Fig.~\ref{fig:ovs_step_by_step} illustrates how different prompt sets affect pseudo-labeling results. When only in-distribution dataset classes, including the ``background'' category, are used ($N_{in}$), OOD objects such as giraffes tend to be misclassified as one of the predefined categories (case (a)). By contrast, incorporating additional OOD-relevant categories into the text prompt set leads to more accurate pseudo-labels (cases (b)--(f)). These results suggest that including candidate classes relevant to potential OOD objects is crucial for improving pseudo-labeling quality and reducing class confusion.

To further verify this observation quantitatively, we performed an experiment where the UniMatch baseline was trained using pseudo-labels derived from cases (a) and (f) in Fig.~\ref{fig:ovs_step_by_step}. As shown in Table \ref{table:ablation_textual_set}, when only in-distribution classes are used as the text prompt set (i.e., case (a)), the additional unlabeled datasets fail to provide any performance benefit, and even result in performance degradation. In contrast, when using a text prompt set that includes half of the full set (i.e., $N$/2), we observe a modest improvement in performance, indicating partial benefits from the OVS predictions. However, the gains remain limited compared to using the full prompt set ($N$). This highlights the importance of covering a broad range of class concepts (including potential OOD categories) to generate more accurate pseudo-labels.
\begin{table}[t]
\caption{Ablation study on the text prompt set in the proposed method. The experiment was conducted on the Pascal VOC setup with $N_{in}$=21 and $N$=171. (a) and (f) correspond to (a) and (f) in Fig.~\ref{fig:ovs_step_by_step}.}
\label{table:ablation_textual_set}
\centering
\resizebox{\linewidth}{!}{%
\begin{tabular}{l|ccccc}
\toprule
Text Prompt Set & 92 & 183 & 366 & 732 & 1464\\
\midrule
Baseline (R-50) & 71.9 & 72.5 & 76.0 & 77.4 & 78.7 \\
\midrule
\multirow{2}{*}{(a) $T=\{T(C_1), ..., T(C_{N_{in}})\}$} & 72.7 & 74.8 & 75.4 & 76.1 & 77.1 \\
& \green{(\small{+0.8})} & \green{(\small{+2.3})} & \red{(\small{-0.6})} & \red{(\small{-1.3})} & \red{(\small{-1.6})}\\
\midrule
\multirow{2}{*}{$T=\{T(C_1), ..., T(C_{N_{in}}), ..., T(C_{N/2}\}$} & 75.3 & 77.3 & 78.1 & 78.7 & 79.0 \\
& \green{(\small{+3.4})} & \green{(\small{+4.8})} & \green{(\small{+2.1})} & \green{(\small{+1.3})} & \green{(\small{+0.3})} \\
\midrule
\multirow{2}{*}{(f) $T=\{T(C_1), ..., T(C_{N_{in}}), ..., T(C_N)\}$} & 78.2 & 78.7 & 79.5 & 80.0 & 80.2 \\
& \green{(\small{+6.3})} & \green{(\small{+6.2})} & \green{(\small{+3.5})} & \green{(\small{+2.6})} & \green{(\small{+1.5})} \\
\midrule
\midrule
Baseline (R-101) & 75.2 & 77.2 & 78.8 &	79.9 &	81.2 \\
\midrule
\multirow{2}{*}{(a) $T=\{T(C_1), ..., T(C_{N_{in}})\}$} & 75.4 & 76.6 & 78.0 & 79.6 & 79.6 \\
& \green{(\small{+0.2})} & \red{(\small{-0.6})} & \red{(\small{-0.8})} & \red{(\small{-0.3})} & \red{(\small{-1.6})}\\
\midrule
\multirow{2}{*}{$T=\{T(C_1), ..., T(C_{N_{in}}), ..., T(C_{N/2}\}$} & 78.3 & 79.6 & 79.9 & 80.8 & 81.2 \\
& \green{(\small{+3.1})} & \green{(\small{+2.4})} & \green{(\small{+1.9})} & \green{(\small{+0.9})} & \green{(\small{+0.0})} \\
\midrule
\multirow{2}{*}{(f) $T=\{T(C_1), ..., T(C_{N_{in}}), ..., T(C_N)\}$} & 80.4 & 81.3 & 81.6 & 81.7 & 81.8 \\
& \green{(\small{+5.2})} & \green{(\small{+4.1})} & \green{(\small{+2.8})} & \green{(\small{+1.8})} & \green{(\small{+0.6})} \\
\bottomrule
\end{tabular}
}
\end{table}

\begin{figure}[t!]
\centering
\includegraphics[width=1.0\linewidth]{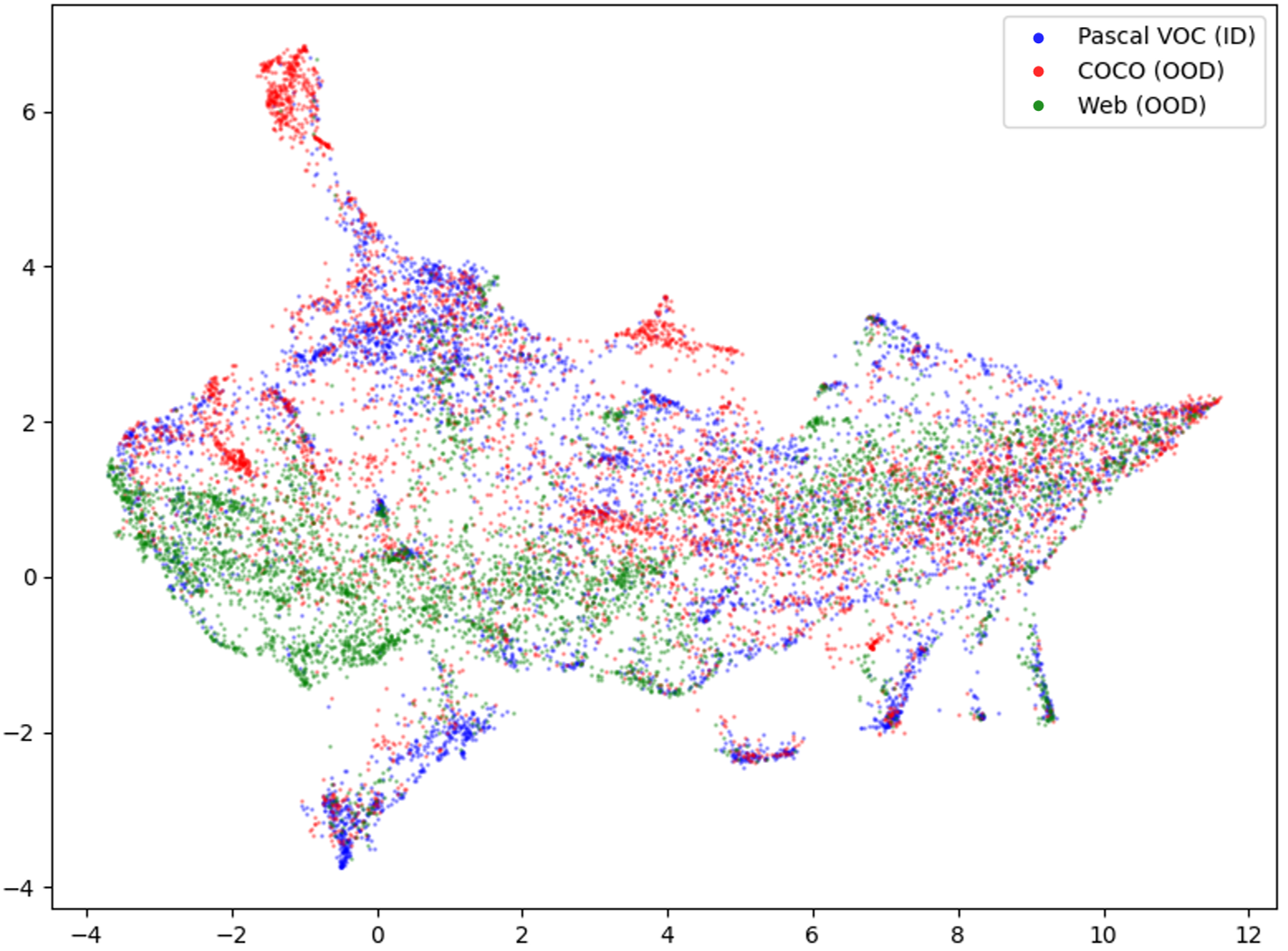}
\caption{UMAP visualization of image representations from three datasets. A total of 15,000 samples were used, with 5,000 randomly selected images from each dataset: Pascal VOC, COCO, and Web-scraped data. Each point represents image features extracted from the encoder and projected into 2D space using UMAP \cite{mcinnes2018umap}.}
\label{fig:umap}
\end{figure}

\subsection{Dataset characteristics and distribution analysis}
\label{subsec:vis_ood}
\begin{figure}[t!]
\centering
\includegraphics[width=1.0\linewidth]{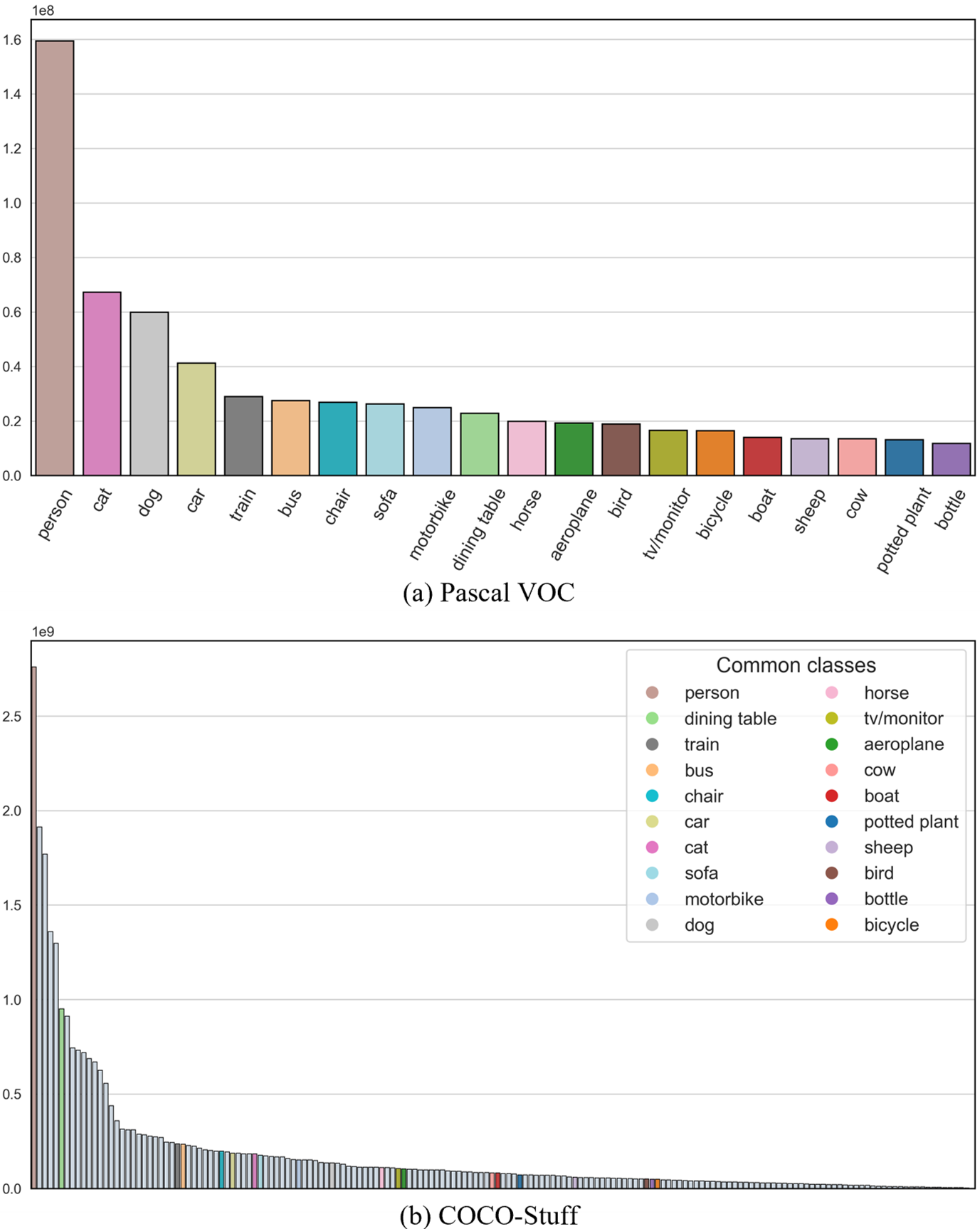}
\caption{Class-wise pixel frequency distributions in segmentation annotations. The x-axis denotes each class, and the y-axis indicates the total number of pixels annotated with that class across the dataset. In COCO-Stuff (i.e., (b)), classes overlapping with Pascal VOC are highlighted in color for comparison, while the remaining classes are shown in gray.}
\label{fig:data_distribution}
\end{figure}

In Fig.~\ref{fig:umap}, we observe that certain regions in the embedding space---such as around (x=–1, y=6), (x=4, y=3), and (x=–2, y=1.5)---are predominantly occupied by COCO samples, suggesting that COCO contains visual concepts and classes that are not present in Pascal VOC. Furthermore, the web-scraped images form distinctive clusters in the regions x=–4 to 4 and y=–1 to 1, differing not only from VOC but also from COCO.
Despite these differences, a large portion of the embedding space exhibits overlap between the datasets, which can be better understood through the class-wise pixel frequency distributions shown in Fig.~\ref{fig:data_distribution}. Notably, person is the most frequent class in both VOC and COCO-Stuff, and 15 out of the 20 Pascal VOC classes (ranging from person to boat in (b)) fall within the top 50\% of the COCO-Stuff class distribution. These observations indicate that, while COCO contains additional visual concepts not in Pascal VOC, the two datasets still share many semantic categories. Consequently, their global-level image representations may appear similar in the embedding space.

Nevertheless, as evidenced by the pseudo-label visualizations in Fig.~\ref{fig:intro}(b) and the quantitative evaluations presented in this paper, pixel-level pseudo-labeling remains significantly more challenging.
Unlike global-level predictions, pixel-wise predictions are more sensitive to class boundaries and local context. Therefore, even subtle differences in visual concepts between in-distribution and OOD images can result in substantial noise in pseudo-labels.
This noise ultimately hampers the training of the semi-supervised segmentation model, limiting the potential benefits of leveraging additional unlabeled datasets. Therefore, to mitigate this problem, we adopt the OVS model, which enables the generation of more robust and reliable pseudo-labels.

\section{Conclusion, Limitations, and Future Work}
\label{Conclusion} 
In this work, we introduced SemiOVS, a novel semi-supervised semantic segmentation framework that effectively leverages unlabeled OOD images via an OVS model.
The two main contributions of this study are: (1) exploring a critical aspect of semi-supervised semantic segmentation by investigating the potential benefits of using larger sets of unlabeled images from readily available sources, such as web-scraped images or images from large-scale datasets, and (2) demonstrating that the OVS model can effectively handle OOD images, leading to significant performance improvements in semi-supervised semantic segmentation.
Extensive experiments on two benchmark datasets show that handling additional unlabeled images via an OVS model can significantly improve performance.
As a result, SemiOVS achieves state-of-the-art performance across different evaluation protocols. It also preserves inference speed because it is a training method that does not affect the inference process. Furthermore, it is compatible with existing semi-supervised segmentation methods, enabling seamless integration into various applications.

The proposed method has shown promising improvements of semi-supervised learners by effectively leveraging additional unlabeled data and pretrained models. However, our evaluation is limited to the natural image domain, where unlabeled data and pretrained models are readily accessible. 
Recently, domain-specific pretrained models have been actively developed to address real-world challenges, such as industrial defect segmentation, medical imaging, and specialized visual content. Therefore, we intend to explore the applicability of our framework to these diverse domains.
Furthermore, the current approach uses web images without additional filtering to maintain practical efficiency. However, this may include diverse image types, such as cartoons or cases where images do not contain target objects, which could potentially limit performance improvement by introducing information that may not benefit the target domain. Therefore, a promising direction for future research would be developing effective sample selection strategies to identify and use images containing meaningful information for the target domain.

\section{Acknowledgements}
This research was supported by the BK21 FOUR funded by the Ministry of Education of Korea and National Research Foundation of Korea. This research was also results of a study on the ``Leaders in INdustry-university Cooperation 3.0'' Project, supported by the Ministry of Education and National Research Foundation of Korea.

\bibliography{paper} 
\bibliographystyle{elsarticle-num}

\appendix
\section{Additional Setups and Results}

\subsection{Web-scraping Queries}
\label{appendix_a}
The search queries and the number of images collected for each class are provided in Table~\ref{web_class_count}.

\begin{table*}[t!]
\caption{Web-scraped images: class-wise queries and counts.}
\centering
\resizebox{0.7\linewidth}{!}{%
\begin{tabular}{cccccc}
\toprule
\textbf{Class} & \textbf{Query} & \textbf{Counts} & \textbf{Class} & \textbf{Query} & \textbf{Counts} \\ \midrule

\multirow{3}{*}{aeroplane} & airplane in airport & 486 & \multirow{2}{*}{cat} & cat in home & 532 \\ 
 & airplane in sky & 554 & & cat in park & 512 \\ 
 & airplane landing image & 463 & & & \\ \midrule
 
\multirow{3}{*}{bicycle} & biking & 600 & \multirow{2}{*}{car} & car in the road & 450 \\ 
 & mountain biking & 561 & & sports car in road & 464 \\ 
 & road biking & 521 & & & \\ \midrule

\multirow{3}{*}{bird} & bird in the sky & 626 & \multirow{3}{*}{horse} & horse in mountains & 505 \\ 
 & bird pictures & 494 & & horse in field & 458 \\ 
 & birds with long legs & 565 & & horse in pasture & 487 \\ \midrule

\multirow{3}{*}{boat} & boat in river & 403 & \multirow{3}{*}{tv} & tv with stand in home & 686 \\ 
 & fishing boat pictures & 376 & & compact tv unit & 632 \\ 
 & floating boat picture & 542 & & tv with desk & 557 \\ \midrule

\multirow{3}{*}{bottle} & bottle on table & 443 & \multirow{3}{*}{monitor} & desk shelf with monitor arm & 669 \\ 
 & bottle with desk & 569 & & monitor in table & 562 \\ 
 & drinking bottle water image & 214 & & monitor on desk & 493 \\ \midrule

\multirow{4}{*}{bus} & bus in the philippines & 463 & \multirow{3}{*}{motorbike} & motorbike in road & 432 \\ 
 & bus in the road & 596 & & motorbike racing & 503 \\ 
 & bus in usa & 577 & & motorbike riding & 504 \\ 
 & public bus in dubai & 572 & & & \\ \midrule

\multirow{4}{*}{chair} & chair with arms in home & 565 & \multirow{4}{*}{person} & us movie still shot & 338 \\ 
 & chairs for bedroom & 598 & & family picture & 505 \\ 
 & chairs for living room & 461 & & instagram selfie & 609 \\ 
 & people sitting on the chair picture & 514 & & real image two people & 250 \\ \midrule

\multirow{3}{*}{cow} & cow in desert & 558 & \multirow{3}{*}{potted plant} & potted plant arrangement ideas & 165 \\ 
 & cow in field & 523 & & potted plant in desk & 483 \\ 
 & cow in pasture & 463 & & potted plant in sunlight & 489 \\ \midrule

\multirow{3}{*}{dining table} & dining table & 622 & \multirow{3}{*}{sheep} & sheep in a herd & 528 \\ 
 & dining table in home & 448 & & sheep in pasture & 484 \\ 
 & dining table in kitchen & 565 & & sheep in the field & 477 \\ \midrule

\multirow{3}{*}{dog} & dog in home & 524 & \multirow{3}{*}{sofa} & sofa in front of window & 671 \\ 
 & dog in park & 503 & & sofa in home & 403 \\ 
 & dog in road & 425 & & sofa in living room & 665 \\ \midrule

\multirow{2}{*}{train} & real train images & 525 \\ 
 & train in mountains & 555 \\ \bottomrule

\end{tabular}}
\label{web_class_count}
\end{table*}

\subsection{Qualitative Results}
\begin{figure*}[hb]
\centering
\includegraphics[width=0.92\linewidth]{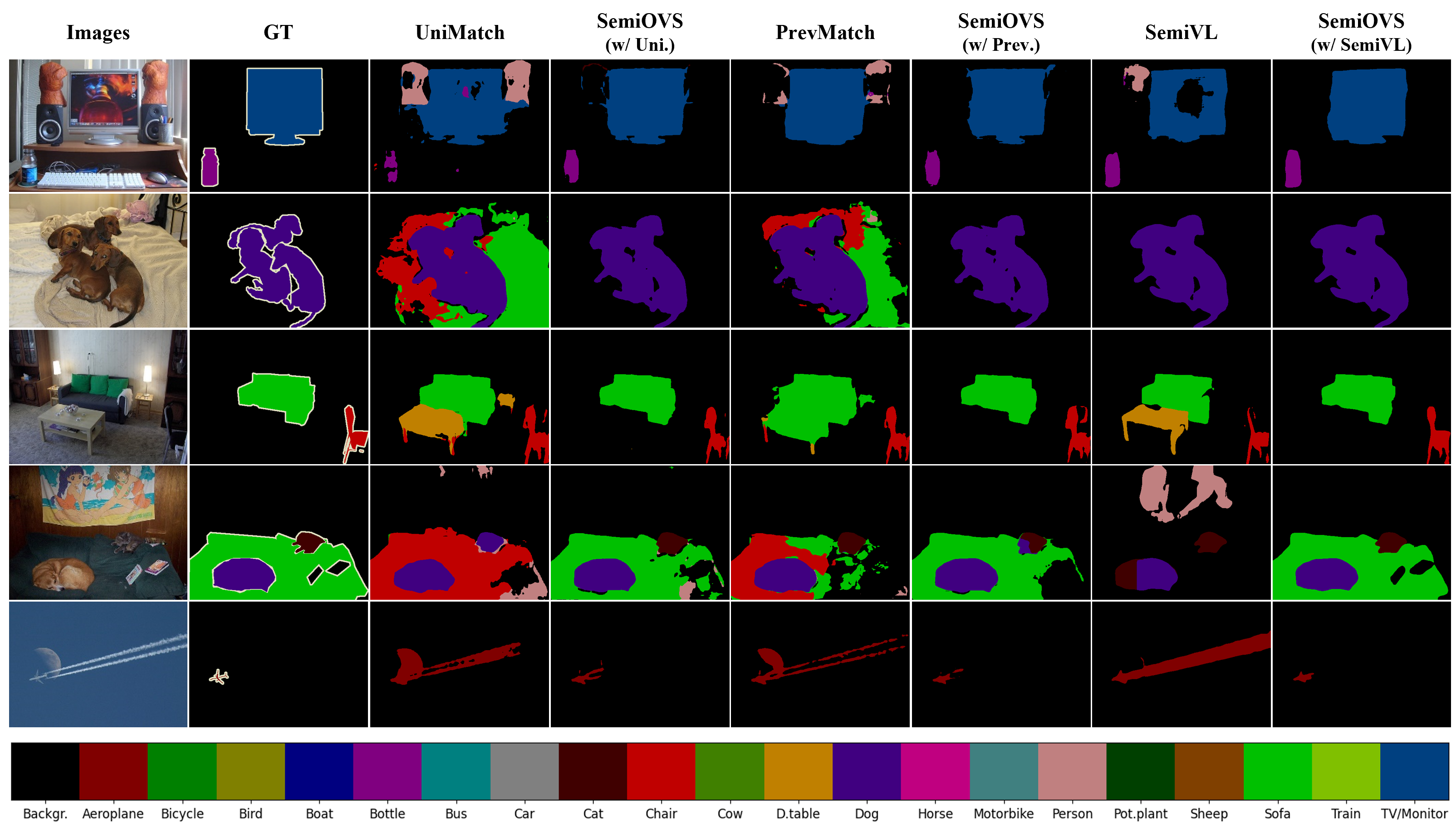}
\caption{Qualitative results on Pascal VOC with a 92-label scenario.}
\label{fig:qualitative}
\end{figure*}

Fig.~\ref{fig:qualitative} presents the qualitative results of the baselines (UniMatch, PrevMatch, SemiVL) and our SemiOVS, revealing that the proposed method more effectively distinguishes classes in complex contextual scenes.

\end{document}